\documentclass[a4paper,10pt,twocolumn]{article}
\usepackage[left=2cm,right=2cm,top=2cm,bottom=2cm]{geometry}

\usepackage{times}

\usepackage{soul}
\usepackage{url}
\usepackage[hidelinks]{hyperref}
\usepackage[utf8]{inputenc}
\usepackage[small]{caption}
\usepackage{graphicx}
\usepackage{amsmath}
\usepackage{booktabs}
\usepackage[numbers]{natbib}
\usepackage{researchpack}
\usepackage{xcolor}
\usepackage{xspace}
\usepackage{booktabs}
\usepackage{multirow}
\usepackage[switch]{lineno}

\newcommand{\method}{TRaCking Knowledge Drift\xspace}
\newcommand{\acronym}{\textsc{trckd}\xspace}
\newcommand{\ACRONYM}{TRCKD\xspace}

\newcommand{\dataset}{\ensuremath{S}}
\newcommand{\distrib}{\ensuremath{P}}

\newcommand{\pastwindow}{\ensuremath{W_\text{old}}\xspace}
\newcommand{\currwindow}{\ensuremath{W_\text{cur}}\xspace}
\newcommand{\MMD}{\ensuremath{\text{MMD}}}
\renewcommand{\cite}{\citep}

\title{Human-in-the-loop Handling of Knowledge Drift}

\author{
Andrea Bontempelli$^1$\footnote{Contact Author}\and
Fausto Giunchiglia$^{1,2}$\and
Andrea Passerini$^1$\and
Stefano Teso$^1$
\\
$^1$University of Trento, Italy, $^2$Jilin University, Changchun, China\\
name.surname@unitn.it
}
\date{}

\begin{document}
\maketitle

\begin{abstract}
    We introduce and study knowledge drift (KD), a complex form of drift that occurs in hierarchical classification.  Under KD the vocabulary of concepts, their individual distributions, and the \emph{is-a} relations between them can all change over time.
    The main challenge is that, since the ground-truth concept hierarchy is unobserved, it is hard to tell apart different forms of KD.  For instance, introducing a new \emph{is-a} relation between two concepts might be confused with individual changes to those concepts, but it is far from equivalent.  Failure to identify the right kind of KD compromises the concept hierarchy used by the classifier, leading to systematic prediction errors.
    Our key observation is that in many human-in-the-loop applications (like smart personal assistants) the user knows whether and what kind of drift occurred recently.
    Motivated by this, we introduce \acronym, a novel approach that combines \emph{automated} drift detection and adaptation with an \emph{interactive} stage in which the user is asked to disambiguate between different kinds of KD.
    In addition, \acronym implements a simple but effective knowledge-aware adaptation strategy.
    Our simulations show that often a handful of queries to the user are enough to substantially improve
    prediction performance on both synthetic and realistic data.
\end{abstract}

\section{Introduction}

We are concerned with human-in-the-loop applications of hierarchical classification under drift.  In such applications, the concept hierarchy embedded into the predictor might become obsolete over time~\cite{stojanovic2002user}.  Smart personal assistants (PAs), for instance, often need to infer the location or social context of their user from sensor data (e.g., GPS coordinates, nearby Bluetooth devices), but the hierarchy of relevant places and people changes as the user's life changes~\cite{giunchiglia2017personal}.  Another example is protein function prediction, in which annotations are structured according to the Gene Ontology, a resource that is regularly updated to incorporate the latest discoveries~\cite{jacobson2018monitoring}.  We refer to this as \emph{knowledge drift} (KD).

Classifiers that fail to adapt to KD may output wrong, ambiguous, or irrelevant predictions.  Standard approaches for learning under concept drift, however, are insufficient for handling KD~\cite{gama2014survey}.  Indeed, KD subsumes but is substantially more complex than concept drift, as it can affect the concept hierarchy itself:  both concepts and \emph{is-a} relations between them may appear, disappear, and change.

The main challenge is how to reliably distinguish between different kinds of KD.  For instance, introducing an \emph{is-a} relation between two concepts leaves a similar footprint on the data stream as changing their individual distributions.  However, confusing one for the other entails acquiring spurious \emph{is-a} relations and hence systematically mis-predicting future instances.

Our key observation is that, in our human-in-the-loop setting, the user can identify with little effort what kind of drift occurred, if any.  Several examples are given below.  Motivated by this observation, we design \acronym (\method), an approach that tackles KD by combining \emph{automated} drift detection and adaptation with \emph{interactive} drift disambiguation.  At a high level, \acronym detects possible KD by checking whether the distribution of current and past examples have diverged using the maximum mean discrepancy~\cite{gretton2012kernel}, and whenever it finds any drift it asks the user to identify what changed.  In order to facilitate this step, \acronym supplies the user with an initial guess and optionally with concrete examples in support of the guess.  Moreover, \acronym implements a simple but effective knowledge-aware adaptation strategy that we ground on top of $k$NN-based multi-label classifiers~\cite{spyromitros2011dealing}.  Our experiments show that interactive drift disambiguation and knowledge-aware adaptation are key for good performance under KD, and that asking a handful of queries to the user is often enough to achieve substantial performance improvements.

\begin{figure*}[tb]
    \centering
    \includegraphics[width=0.95\linewidth]{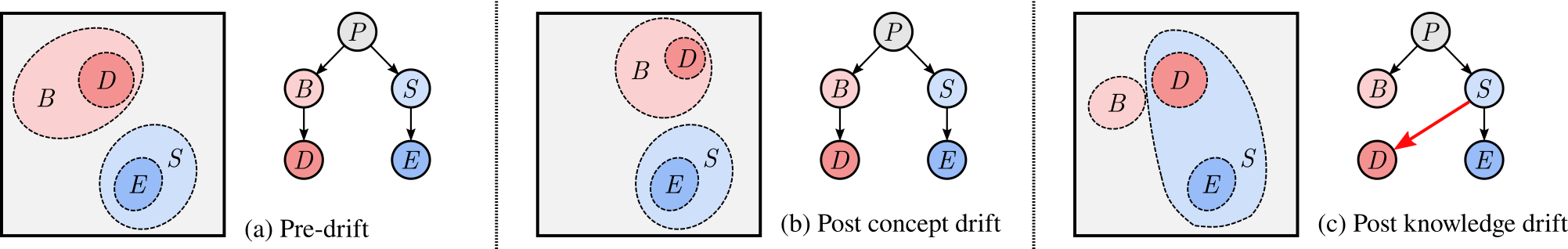}
    \caption{
    \textbf{Left}:  Decision surface and concept hierarchy of a classifier for Ann's social context that knows about five concepts: ``\emph{P}erson'', ``\emph{B}oss'', ``\emph{S}ubordinate'', ``\emph{D}ave'', and ``\emph{E}arl''.
    \textbf{Middle}:  Concept Drift:  Dave moves to a different office.  As new examples are received, the decision surface changes but the hierarchy remains the same.
    \textbf{Right}:  Knowledge Drift:  Ann is promoted and ``Dave'' is now her subordinate.  If the classifier knows that the hierarchy changed, it can transfer examples from ``Dave'' to ``Subordinate'', quickly improving its performance.}
    \label{fig:example}
\end{figure*}

\paragraph{Contributions:}  Summarizing, we:
\begin{enumerate}

    \item Introduce knowledge drift and the related issue of drift disambiguation.

    \item Design \acronym, an approach to KD in human-in-the-loop applications that combines \emph{automated} detection and adaptation with \emph{interactive} disambiguation, and instantiate it on top of $k$NN-based approaches to concept drift.

    \item Show experimentally that, thanks to knowledge-aware adaptation guided by few user queries, \acronym often outperforms its competitors on three data sets.
    
\end{enumerate}

\section{Preliminaries}

\noindent We begin by introducing hierarchical classification under drift.  Instances $\vx$ belong to one or more concepts (\emph{aka} classes) organized in a ground-truth hierarchy, a DAG in which nodes are concepts and edges encode \emph{is-a} relations.  Labels $\vy$ are indicator vectors:  the $i$th element of $\vy$, denoted $y^i$, is $1$ if $\vx$ belongs to the $i$th concept and $0$ otherwise.

\begin{example}
    Given observations $\vx$ (e.g., GPS coordinates and list of nearby devices), Ann's PA predicts that Ann is studying at the library with Bob.  Here, ``Studying'', ``Library'', and ``Bob'' are concepts, and the concept hierarchy states, among other things, that ``Bob'' is a ``Friend'' and a ``Person''.
\end{example}

\noindent The machine observes a stream of examples $\vz_t = (\vx_t, \vy_t)$ drawn from a ground-truth data distribution $\distrib_t(\vX, \vY)$.  The latter is always consistent with the ground-truth hierarchy: if the $j$th concept \emph{is-a} specialization of the $i$th concept, then $y^j = 1$ implies $y^i = 1$ (and conversely $y^i = 0$ implies $y^j = 0$).  The machine's goal is to \emph{learn a classifier that performs well on future instances}.\footnote{We assume $\vy$ to be given for ease of exposition.  In practical applications, often $\vy$ has to be acquired using an active learning step.}

\subsection{Knowledge Drift}

What makes this setting challenging is that the ground-truth concept hierarchy and the data distribution can both change over time $t$ and that neither of them can be observed directly.  This gives rise to a very general form of drift that we denote \emph{knowledge drift} (KD).

In standard (i.e., non-hierarchical, closed world) settings, the only possible forms of drift are \emph{distribution shift}, in which the prior distribution over instances $\distrib_t(\vX)$ changes, and \emph{individual concept drift}, in which the conditional distribution of a single concept $\distrib_t(Y_i\,|\,\vX)$ changes~\cite{gama2014survey}.  The term \emph{concept drift} captures both processes.

\begin{example}
During the semester, Ann spends most of her time studying at the library.  Once the finals are over, Ann stops going to the library as often and while there she is less likely to be studying.  This affects the distribution of GPS coordinates and that of activities conditioned on GPS coordinates.
\end{example}

\noindent  Knowledge drift subsumes, but is more complex than, concept drift, as it can affect the concept hierarchy itself.  Four atomic types of drift to the hierarchy can be defined:  \emph{concept addition}, \emph{concept removal}, \emph{relation addition}, and \emph{relation removal}.   Concept addition refers to the appearance of new concepts and concept removal to the phasing out of obsolete concepts.  Relation addition and relation removal, on the other hand, refer to changes in the structure of the hierarchy itself.

\begin{example}
Concepts like ``Friend'' and ``Library'' are essentially immutable, but the specific friends and libraries that matter to Ann (which are also concepts) can change over time, e.g., when Ann graduates or moves abroad.
In particular:
i)~If Ann buys a vacation home, an unanticipated concept ``Ann's vacation home'' appears in the ground-truth hierarchy (concept addition).  Conversely, if Ann's vacation home is sold, the corresponding concept is no longer meaningful and disappears (concept removal).
ii)~If Ann receives a promotion and her old boss Dave becomes her subordinate, then ``Dave'' moves from being a child of ``Boss'' (relation removal) to being a child of ``Subordinate'' (relation addition).
\end{example}

\subsection{Handling Knowledge Drift}

Roughly speaking, handling concept drift involves two steps:  \emph{detecting} drift and \emph{adapting} the model accordingly~\cite{gama2014survey}.  Timely detection and adaptation are a prerequisite for avoiding prediction mistakes, even more so in the hierarchical case, where a single misaligned concept can compromise the predictions of all related concepts.

Dealing with KD, however, requires the machine to maintain a copy of the (unobserved) ground-truth concept hierarchy and to update it based on the identified KD.  Doing so involves one extra step:  \emph{drift disambiguation}.  By this we mean identifying what concepts and relations were affected by drift and how.  This step is critical, because different types of KD require different types of adaptation and failure to disambiguate KD can induce a mismatch between the ground-truth and the classifier's hierarchies, leading to systematic, cascading prediction errors~\cite{koller2009probabilistic}.

Solving disambiguation is far from trivial.
To see this, consider relation addition.  Like all other forms of KD, relation addition can only be identified by its effects on the data distribution.  Specifically, adding a relation introduces a \emph{correlation} between the concepts appearing in the relation.  Naturally, the converse does not necessarily hold:

\begin{example}
    Ann is on a work trip and sleeps in her hotel room by herself.  Hence, the concepts ``Sleeping'' and ``Alone'' are highly correlated.  Once Ann gets back home, she no longer sleeps alone and the correlation drops dramatically.
\end{example}

\noindent  This shows that logically independent concepts can be highly correlated.  Now, the machine has no sure way of telling apart relation addition from concept drift, and might react by adding a spurious relation to its concept hierarchy.  It might take plenty of examples for the machine to correct its mistake, and reliable examples are usually scarce in non-stationary settings like ours.  Relation removal works similarly.

Dealing with concept removal is also non-trivial: removing the $i$th concept implies not only that it cannot occur (and should not be predicted) ever again, but also that its child concepts are not longer attached to it.  Treating concept removal as concept drift makes it hard to force the conditional distribution $\distrib_t(Y_i\,|\vX)$ to exactly zero and, depending on the classifier being used, even doing so may not guarantee that the deleted concept is not predicted when one of its children is.\footnote{The only ``easy'' case is concept addition, which is straightforward in our fully labeled setting and will not be considered further.
}

\section{Handling Knowledge Drift with \ACRONYM}

Our approach stems from the simple observation that in many human-in-the-loop scenarios \emph{the user can naturally disambiguate between different types of KD}.  Consider our running example:  Ann is perfectly aware that the finals are not yet over (so no drift occurred in this regard) and that her vacation home has been recently sold (and hence that ``vacation home'' is no longer a valid concept).  It is therefore sensible to partially offload drift disambiguation to the user.

To this end, we introduce \acronym, a $k$NN-based approach for human-in-the-loop hierarchical classification under KD that combines \emph{automated} detection and adaptation with \emph{interactive} disambiguation.  \acronym detects drift using a time-tested sliding-window approach~\cite{kifer2004detecting} upgraded to hierarchical classification.  Most importantly, \acronym introduces an new interactive drift disambiguation step and a simple but effective knowledge-aware adaptation strategy.  Following the literature, we focus on multi-window $k$NN-based classifiers~\cite{spyromitros2011dealing}, however the overall algorithm of can be adapted to more complex models.

The pseudo-code of \acronym is listed in Algorithm~\ref{alg:method}.  The algorithm takes a data set $S_1$ and a concept hierarchy compatible with it and uses them to train an initial classifier.  Then, in each iteration $t = 1, 2, \ldots$ the machine receives a new example $\vz_t = (\vx_t, \vy_t)$ and performs three steps:  1)~It detects whether KD occurred,  2)~It cooperates with the user to determine what concepts and relations were affected by KD, and 3)~It adapts the classifier and the machine's concept hierarchy accordingly.  We discuss these three steps in turn.

\paragraph{Step 1:  Detection.}  \acronym builds on the multi-label two-window detector of~\citet{spyromitros2011dealing}.
For every concept $i$ in the machine's hierarchy, \acronym maintains two windows of examples:\footnote{In hierarchical classification, most examples do not belong to most concepts.  As in~\cite{spyromitros2011dealing}, \acronym accounts for this by allocating $2/3$ of each window for negatives.}  $\currwindow^i$ holds the $w$ most recent examples and is updated in each iteration, while $\pastwindow^i$ holds $w$ reference (past) examples and is only updated when the $i$th concept drifts.  Whenever this happens, the contents of $\pastwindow^i$ are replaced with those of $\currwindow^i$ and the latter is emptied.  Predictions are made using a standard $k$NN rule using only the examples in the recent window.  KD is detected whenever the measured difference between the distributions of the recent and past windows of at least one concept is larger than some threshold $\tau$.

\acronym measures this difference using the \emph{maximum mean discrepancy} (MMD), a well-known discrepancy employed in hypothesis testing~\cite{gretton2012kernel} and domain adaptation~\cite{zhang2013domain}.
Letting $P$ and $Q$ be distributions over some space $\calX$ and $k$ a user-defined kernel over $\calX$, the MMD between $P$ and $Q$ relative to $k$ is:
$$
    \MMD(P,Q)^2 = \bbE[k(\va,\va')] - 2 \bbE[k(\va,\vb)] + \bbE[k(\vb,\vb')]
$$
where $\va$, $\va'$ are drawn i.i.d. from $P$ and $\vb$, $\vb'$ from $Q$.
Estimating the MMD between $\pastwindow^i$ and $\currwindow^i$ requires to define a kernel between examples $\vz$.  \acronym achieves this by defining two separate kernels over instances and labels $k_X$ and $k_Y$ and then taking their tensor product~\cite{srinivas2010gaussian}:
$
    k(\vz,\vz') = k((\vx,y),(\vx',y')) = k_X(\vx,\vx') \cdot k_Y(y,y')
$.
The choice of base kernels is application-specific.  In our experiments, we employ a Gaussian kernel $k_X$ for the instances and a delta kernel $k_Y(y,y') = \Ind{y = y'}$ for the labels.

The MMD has several advantages.  First of all, if $P \equiv Q$ then necessarily $\MMD(P,Q) = 0$.\footnote{If $k$ is characteristic, the converse also holds~\cite{szabo2017characteristic}.}  Unlike other theoretically well-behaved discrepancies (like total variation distance and $\calA$-distance~\cite{kifer2004detecting}), the MMD can be estimated efficiently (in linear or quadratic time, depending on the estimator) even for higher-dimensional data~\cite{paninski2003estimation,perez2009estimation,gretton2012kernel}.  Just as importantly, in our experiments the MMD achieved better false detection rate than ME~\cite{jitkrittum2016interpretable}, a state-of-the-art discrepancy with better discrimination power on paper.

\begin{algorithm}[tb]
    \caption{The \acronym algorithm.
    Inputs:  $\dataset_1$ is the initial data set, $s := |\dataset_1|$, $w$ the window size, $\tau$ a threshold.
    $\widehat{\mathrm{MMD}}$ is an empirical estimator of the MMD and $\vz_t^i := (\vx_t, y_t^i)$.}
    \label{alg:method}
    \begin{algorithmic}[1]
        \State Fit initial classifier on $\dataset_1$
        \For{every concept $i$ in the machine's hierarchy}
            \State $\pastwindow^i \gets \{ \vz_s^i, \ldots, \vz_{s - w}^i \}$ \label{line:init-past-window}
        \EndFor
        \For{$t = 1, 2, \ldots$}
            \State Receive new example $\vz_t$
            \For{every concept $i$ in the machine's hierarchy}
                \State $\currwindow^i \gets \{\vz_{s + t}^i, \vz_{s + t  - 1}^i, \ldots, \vz_{s + t - w}^i\}$ \label{line:update-curr-window}
            \EndFor
            \If{$\exists i \,:\, \widehat{\mathrm{MMD}}(\currwindow^i, \pastwindow^i) \ge \tau$} \label{line:detect-drift}
                \State Illustrate detected KD to the user \label{line:ask-user}
                \State Adapt based on user's KD description \label{line:receive-delta}
            \EndIf
        \EndFor
    \end{algorithmic}
\end{algorithm}

\paragraph{Step 2:  Disambiguation.}   Upon detecting drift, \acronym initiates interaction with the user.  Interaction can be carried out through a simple UI akin to those used for designing ontologies~\cite{jimenez2012large,angeli2015leveraging}.  The interface shows (part of) the machine's concept hierarchy and highlights the concepts that were identified by MMD.  Notice that this visual description of the detected drift is unstructured, in that it does not disambiguate between different forms of atomic KD.
In our running example, the PA might highlight the concepts ``Dave'', ``Boss'', and ``Subordinate'', thus enabling Ann to focus her recent promotion and identify related changes.
Of course, the machine's description of drift might be inaccurate or incomplete.  For this reason, \acronym asks the user to correct it using the graphical interface.
Extra context can be supplied to the user by including a handful of examples that summarize how the concepts affected by KD have changed.  Such examples can be selected from the past and current windows of those concepts using MMD witness functions~\cite{lloyd2015statistical}.

If the machine's drift description is accurate enough (which should typically be if the drift detector is tuned well) then the user's jobs is simply to tell the machine whether the relations between the highlighted concepts have undergone drift and how.  To this end, the user can deselect highlighted concepts and add or remove arrows in the DAG.  This is not particularly cognitively demanding.  If the machine's description is not accurate, the user can still easily deselect wrongly highlighted concepts.  This is already an improvement over no interaction.  A sufficiently motivated and knowledgeable user has also the option of editing any drifting concepts or relations not detected by the machine.  All of this extra information is useful for guiding knowledge-aware adaptation.

\paragraph{Step 3:  Adaptation.}  Once it receives the user's drift description, \acronym adapts the machine's hierarchy and the windows accordingly.  In particular:
i)~For every instance of individual concept drift in the description, it empties the current window of the affected concept and transfers its contents to the past window.
ii)~For concept removal, the past and current windows of the affected are deleted and any \emph{is-a} relations between the removed concept are deleted and its children are attached to the parent.
iii)~For relation addition, the positive examples belonging to the child concept are copied to the parent's window and the latter is doubled in size to cope.
iv)~For relation removal, the positive examples belonging to the child concept are removed from the parent's window and the latter is shrank accordingly.  The child concept is also linked directly to its grand-parent.
This strategy applies immediately to $k$NN-based approaches and to other classifiers that support forgetting (i.e., removal of training examples).
Our experiments show that this simple form of knowledge-aware adaptation strategies outperforms strategies based on adaptively forgetting obsolete examples~\cite{spyromitros2011dealing,roseberry2019multi}.

\begin{figure*}[tb]
    \centering
    \begin{tabular}{ccc}
        \includegraphics[width=0.3\textwidth]{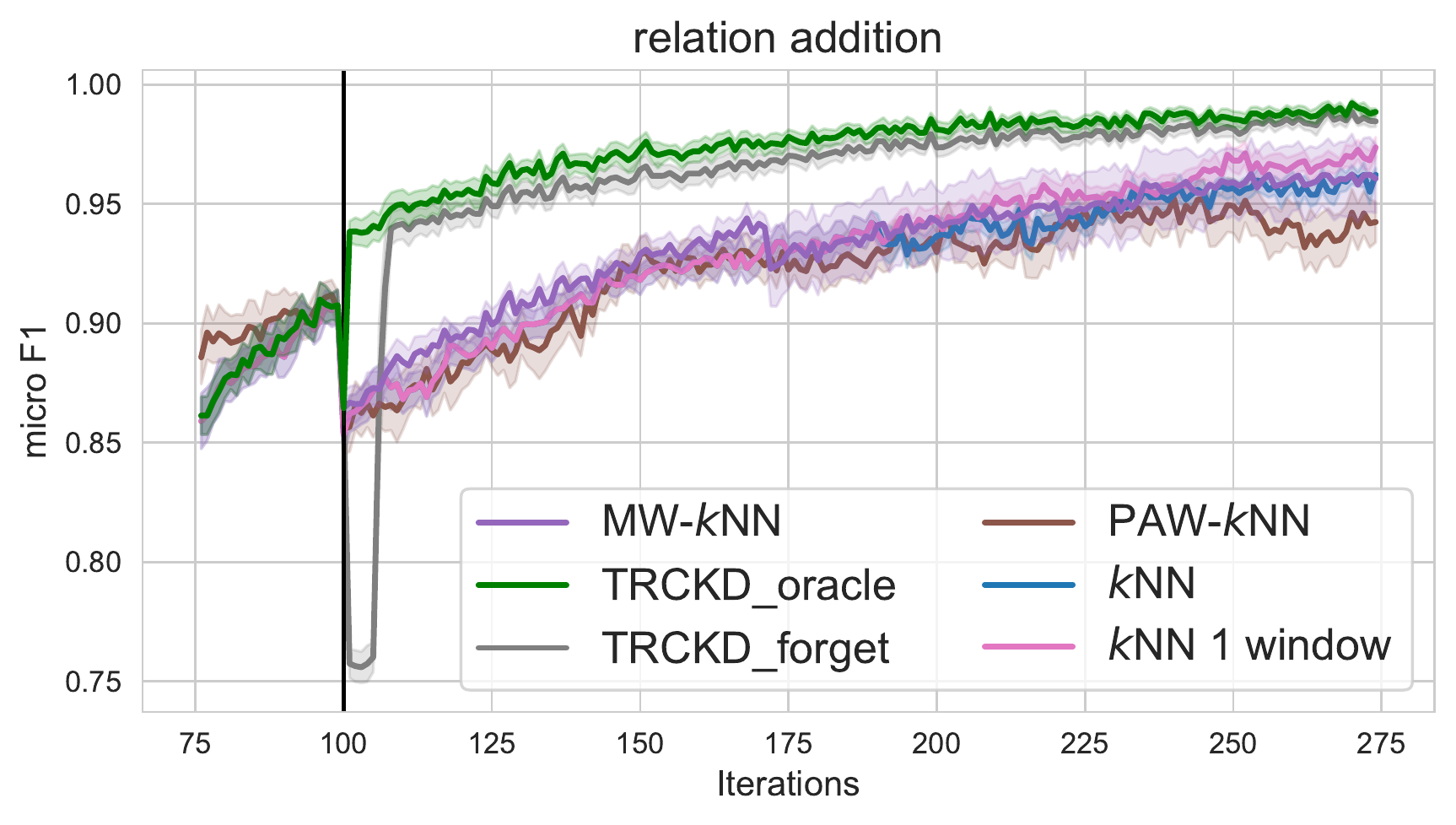} &
        \includegraphics[width=0.3\textwidth]{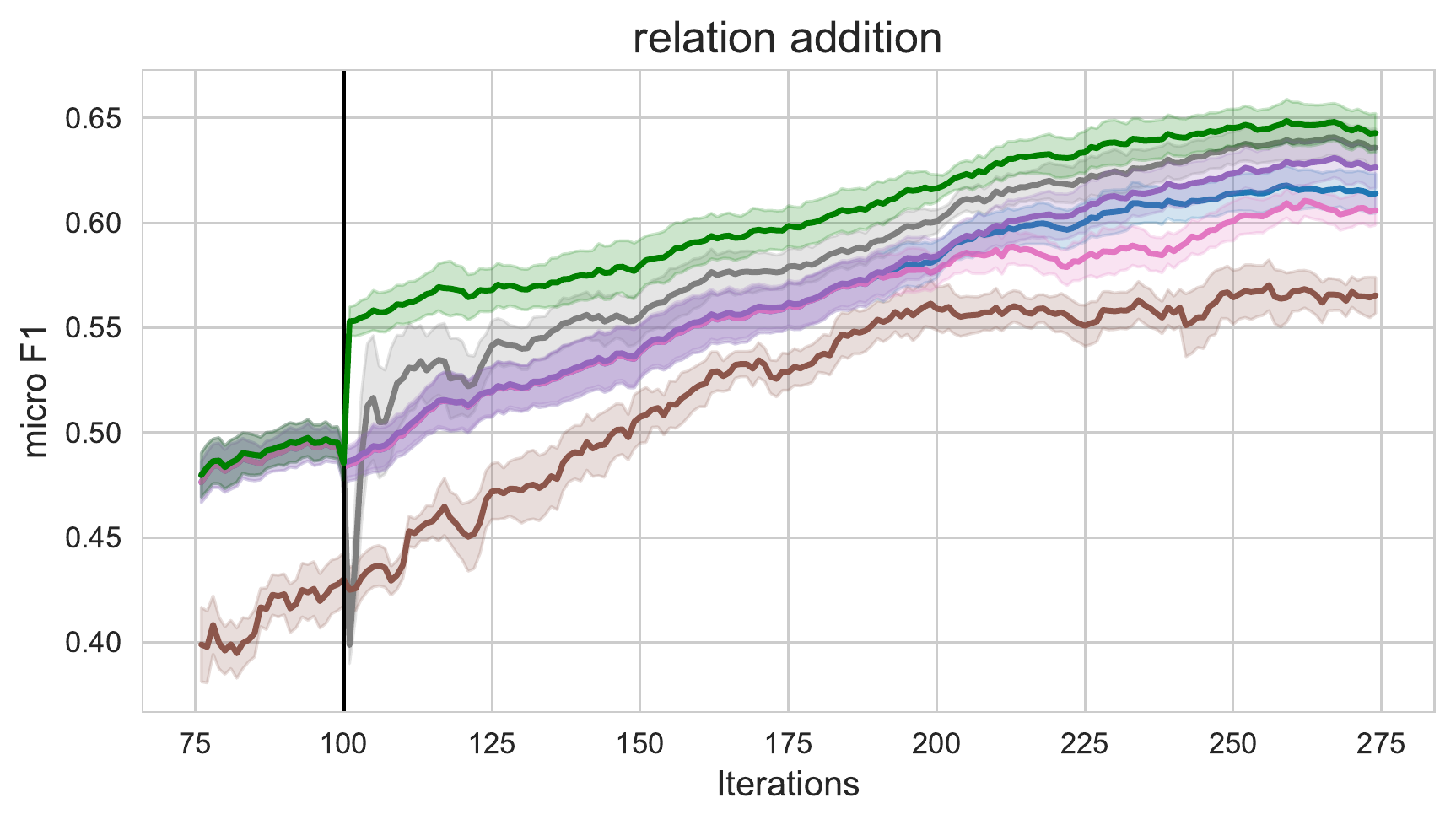} &
        \includegraphics[width=0.3\textwidth]{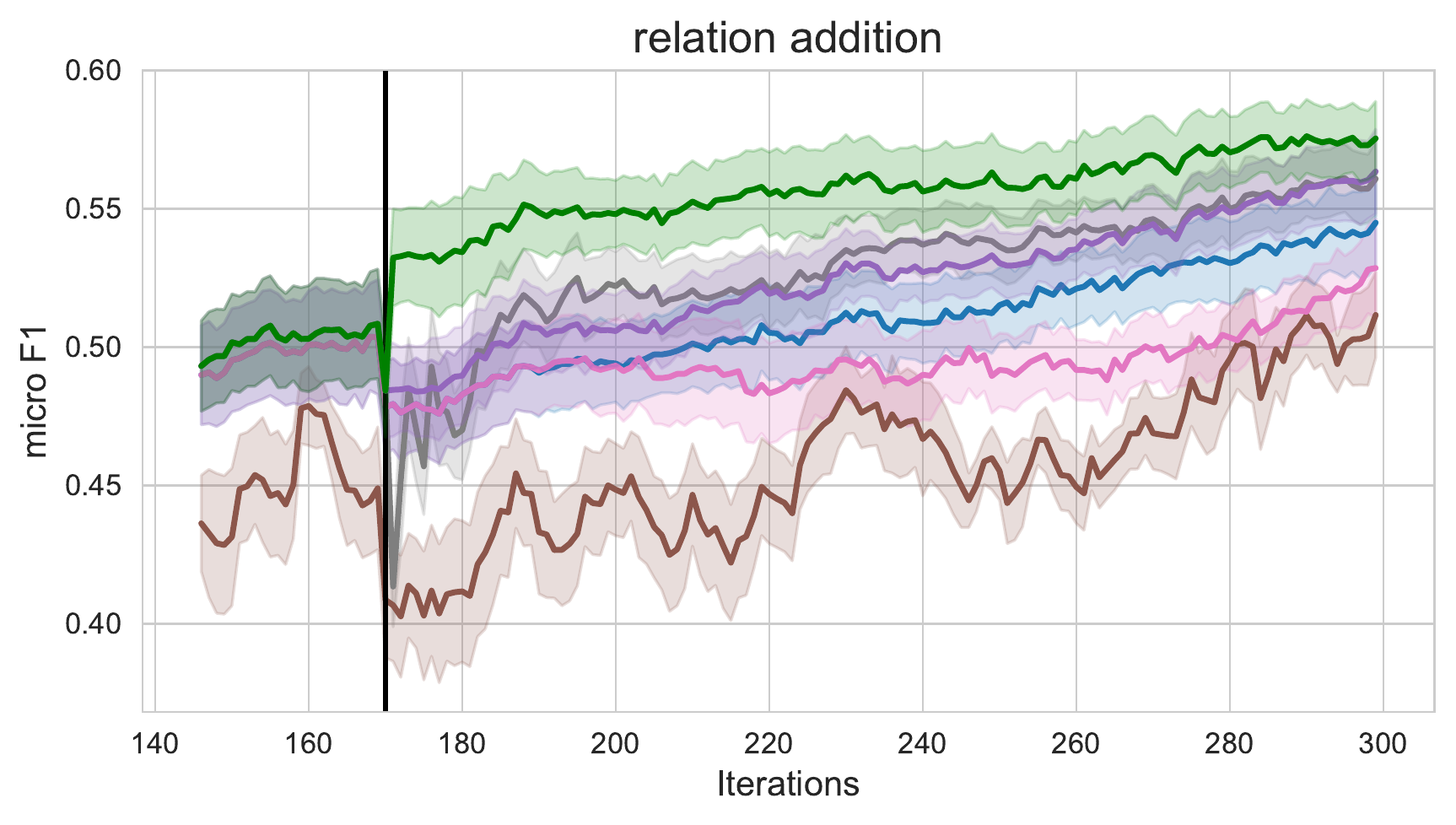}
        \\

        \includegraphics[width=0.3\textwidth]{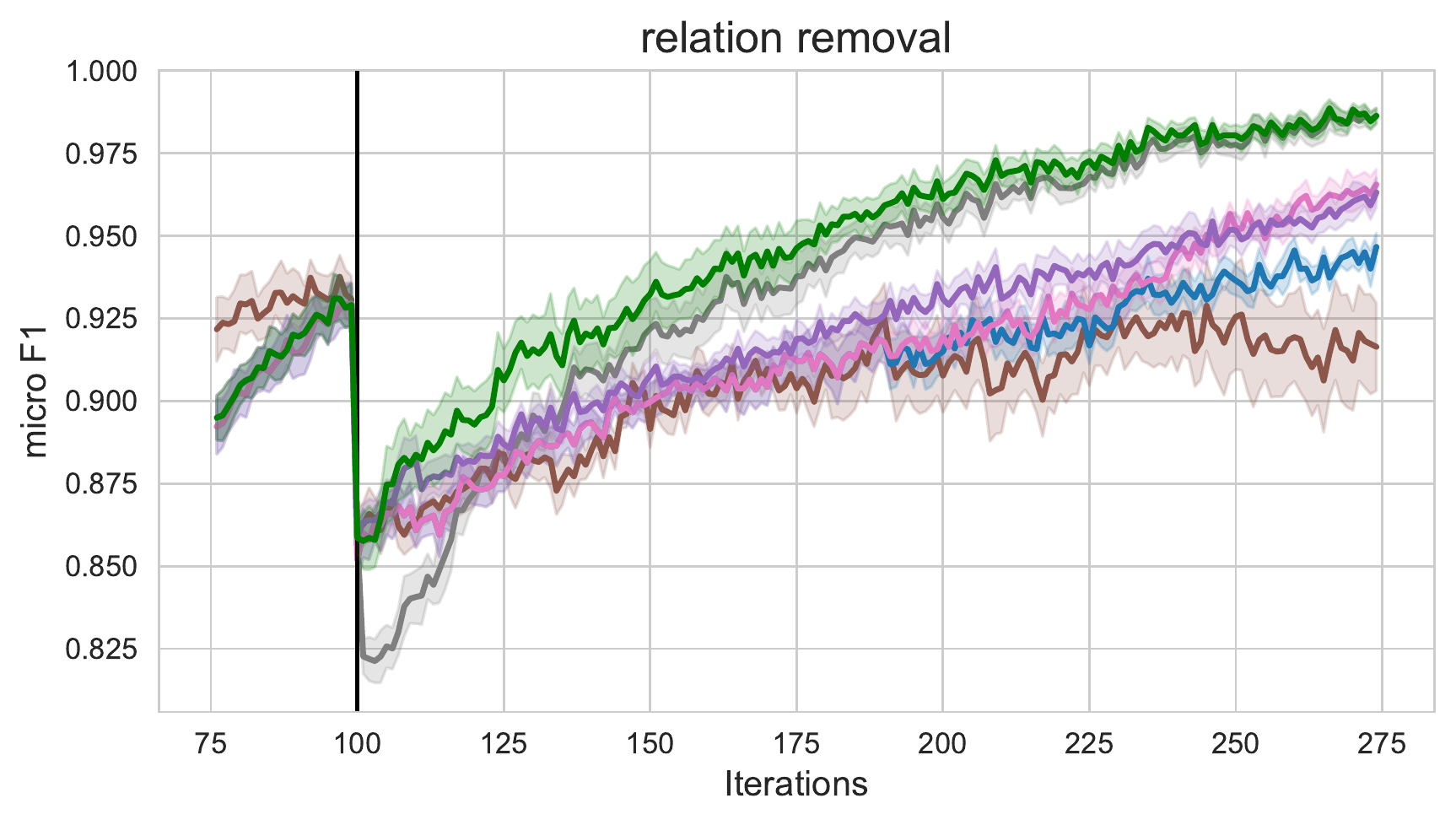} &
        \includegraphics[width=0.3\textwidth]{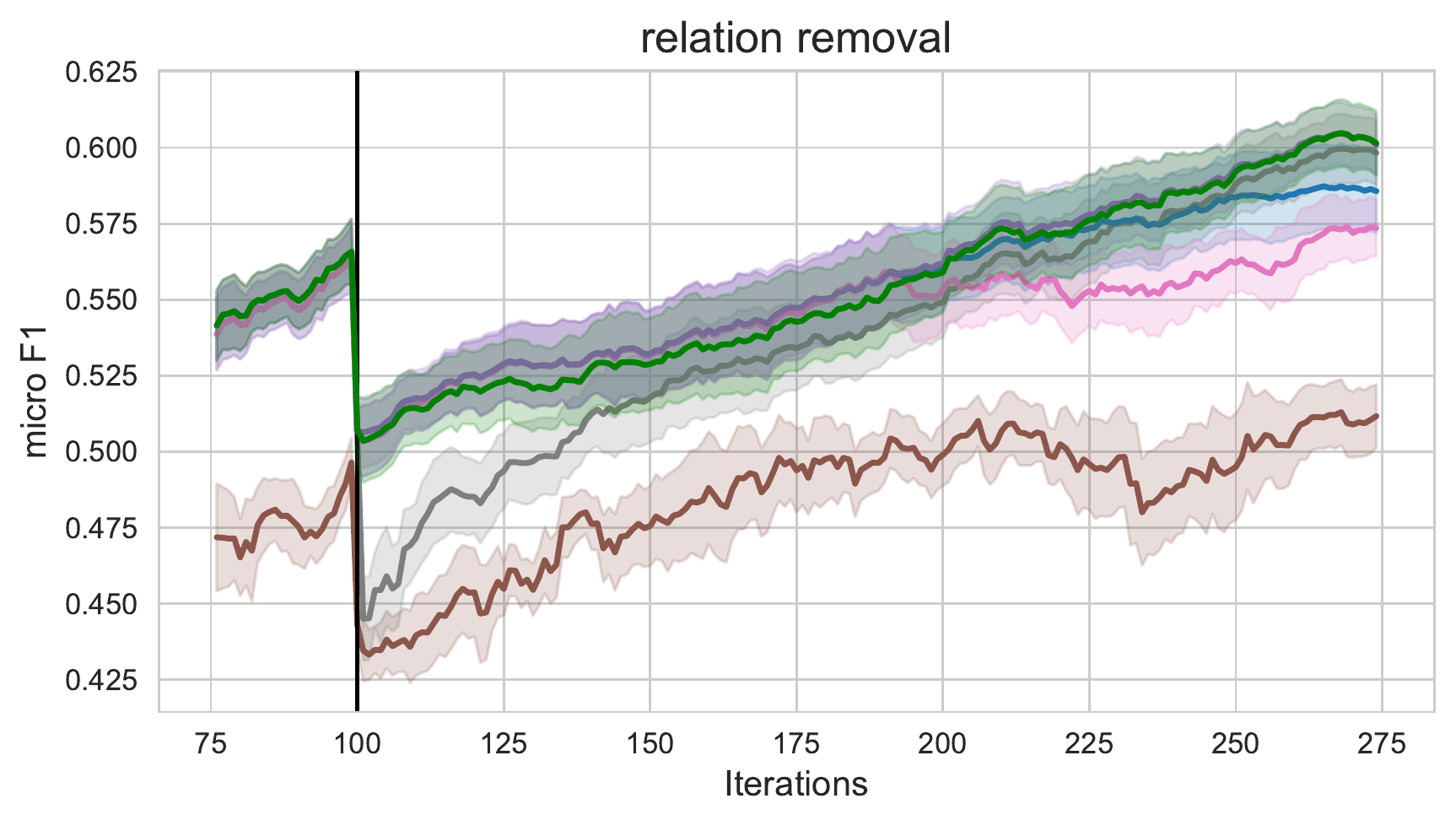} &
        \includegraphics[width=0.3\textwidth]{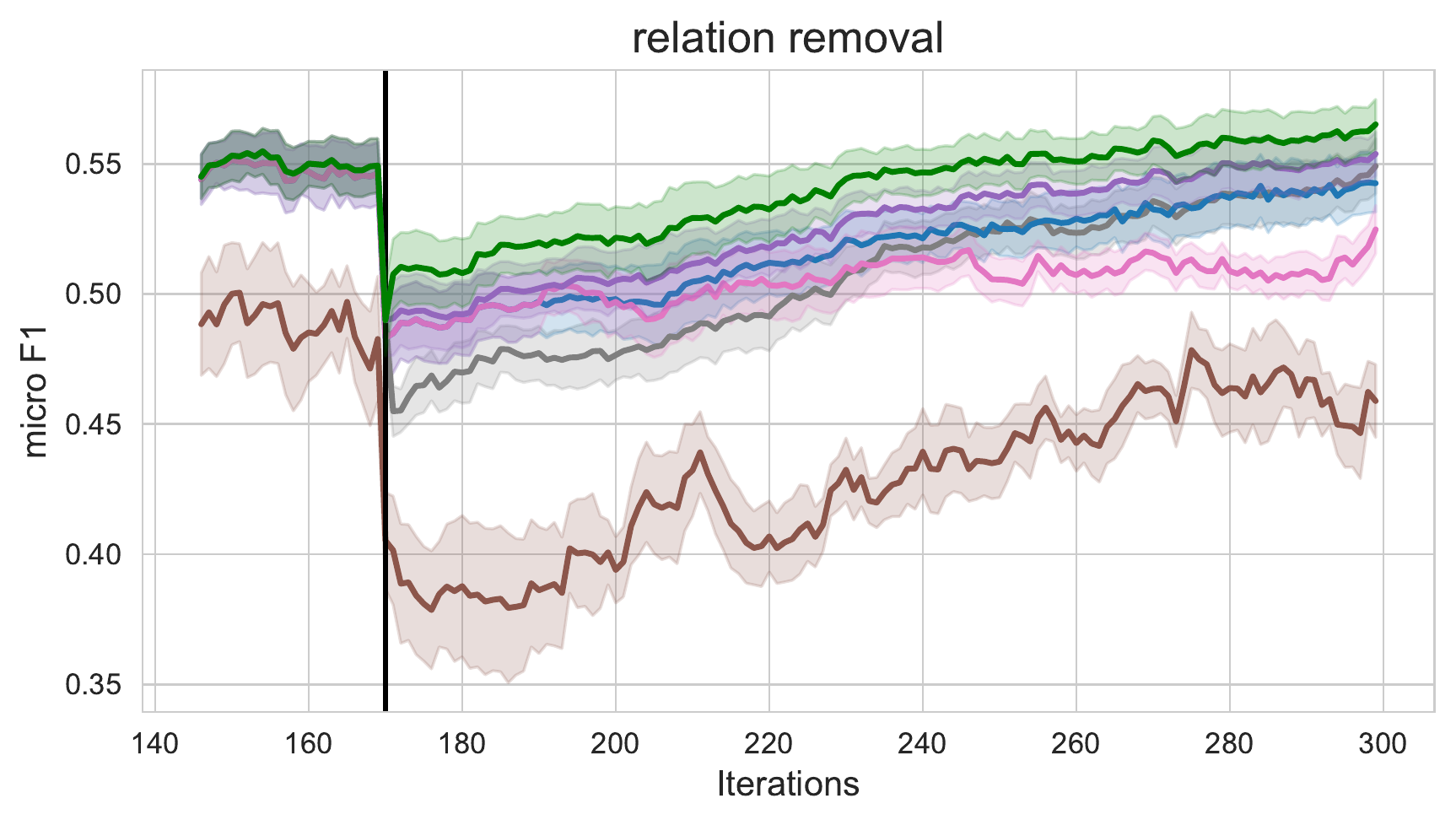}
    \end{tabular}
    
    \caption{Knowledge-aware versus other adaptation strategies.  \textbf{Left to right}:  Results for HSTAGGER, EMNIST, and 20NG.  \textbf{Top row}:  relation addition.  \textbf{Bottom row}:  relation removal.  Plots for the other forms of KD can be found in the Supplementary Material.}
    \label{fig:q1}
\end{figure*}

\begin{figure*}[tb]
    \centering
    \includegraphics[width=0.3\textwidth]{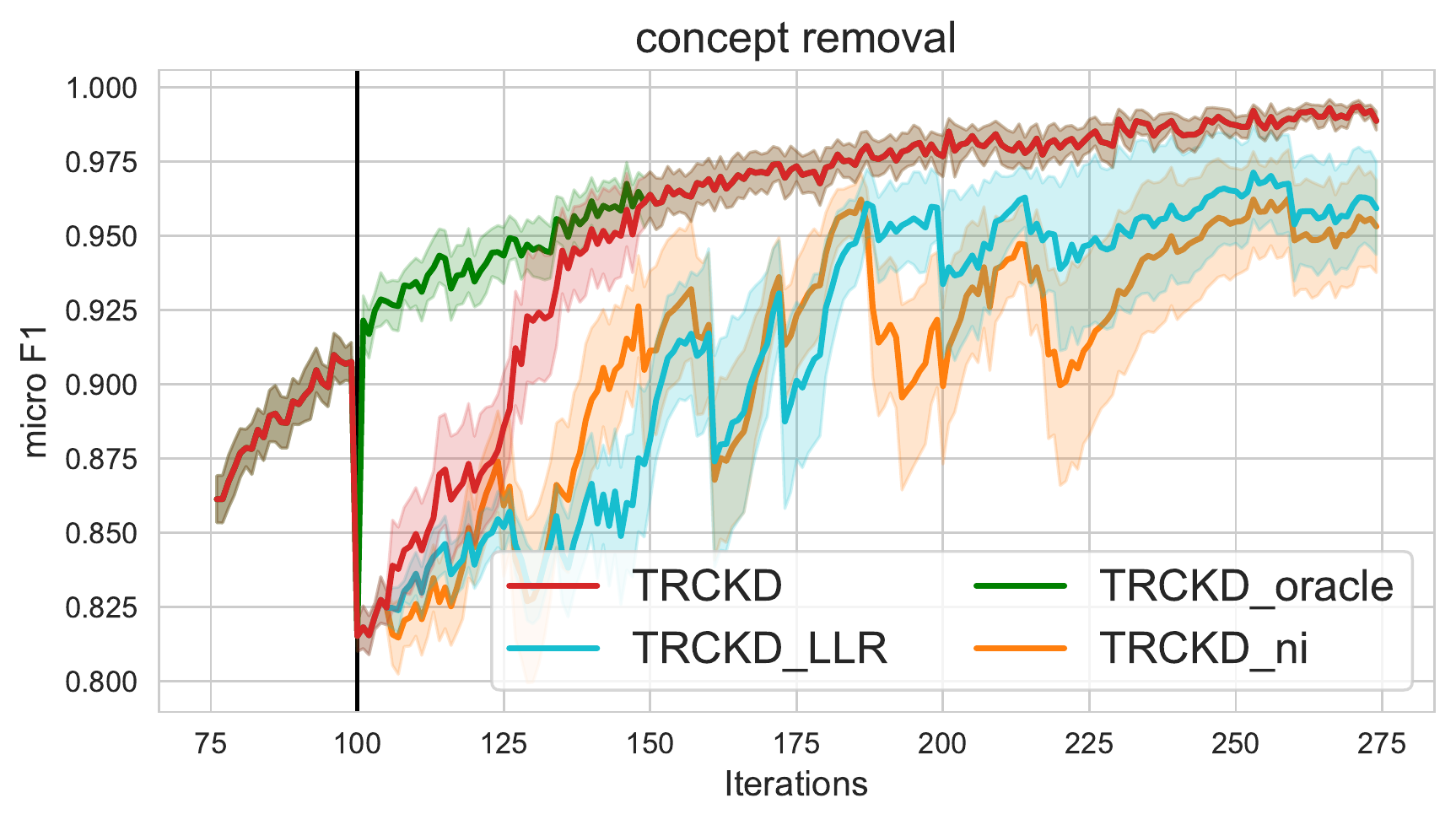}
    \includegraphics[width=0.3\textwidth]{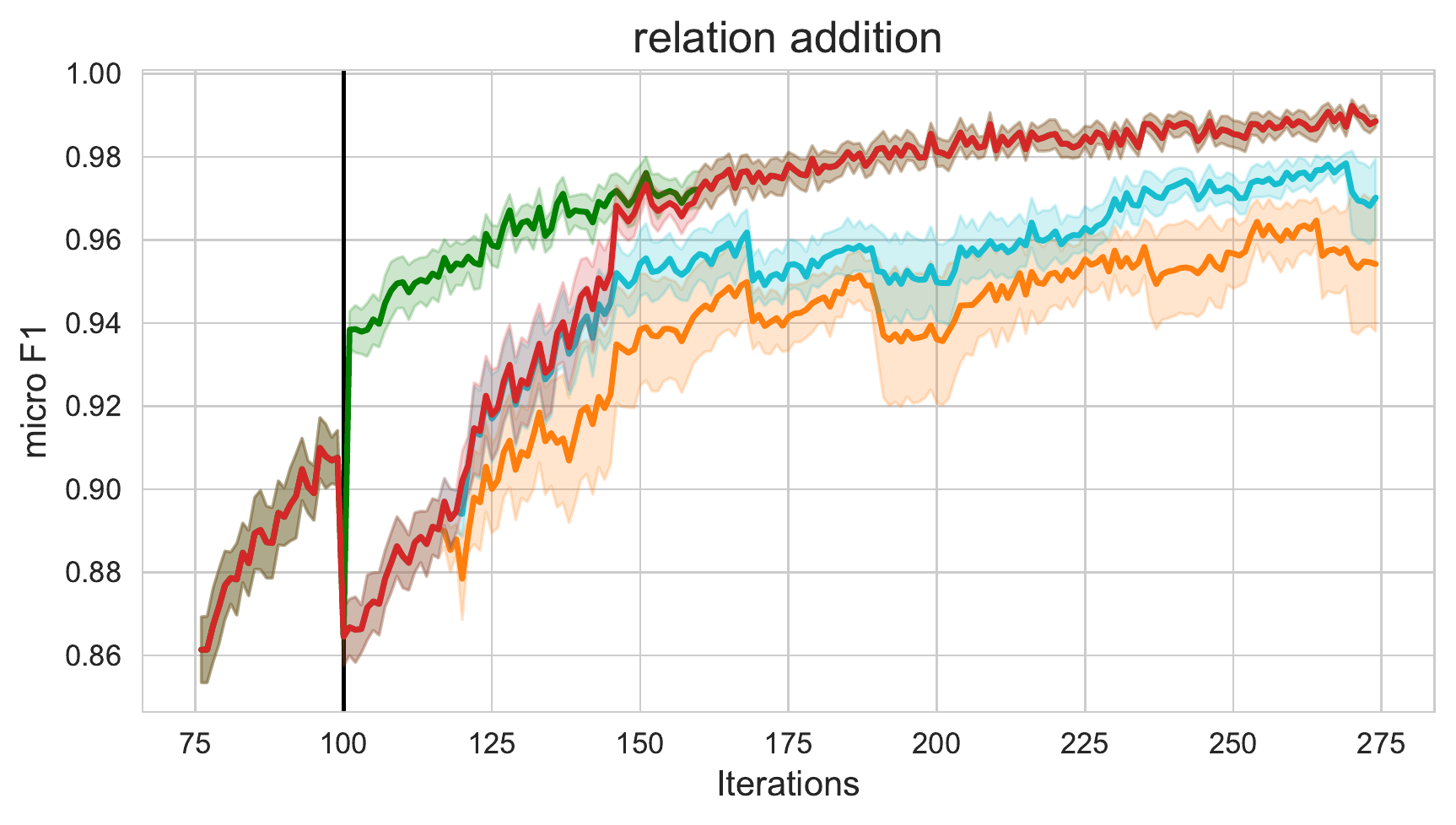}
    \includegraphics[width=0.3\textwidth]{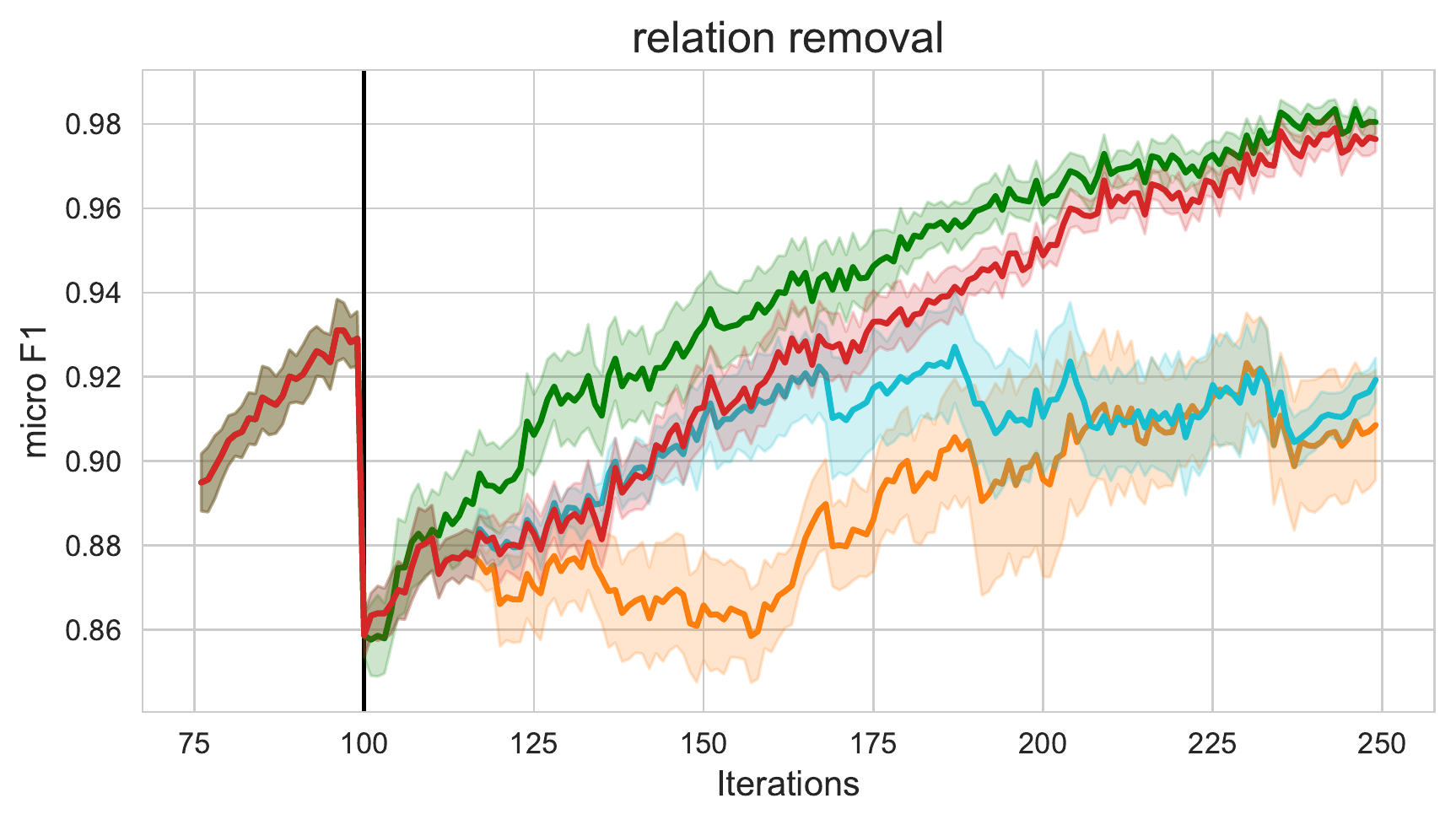} \\
    
    \includegraphics[width=0.3\textwidth]{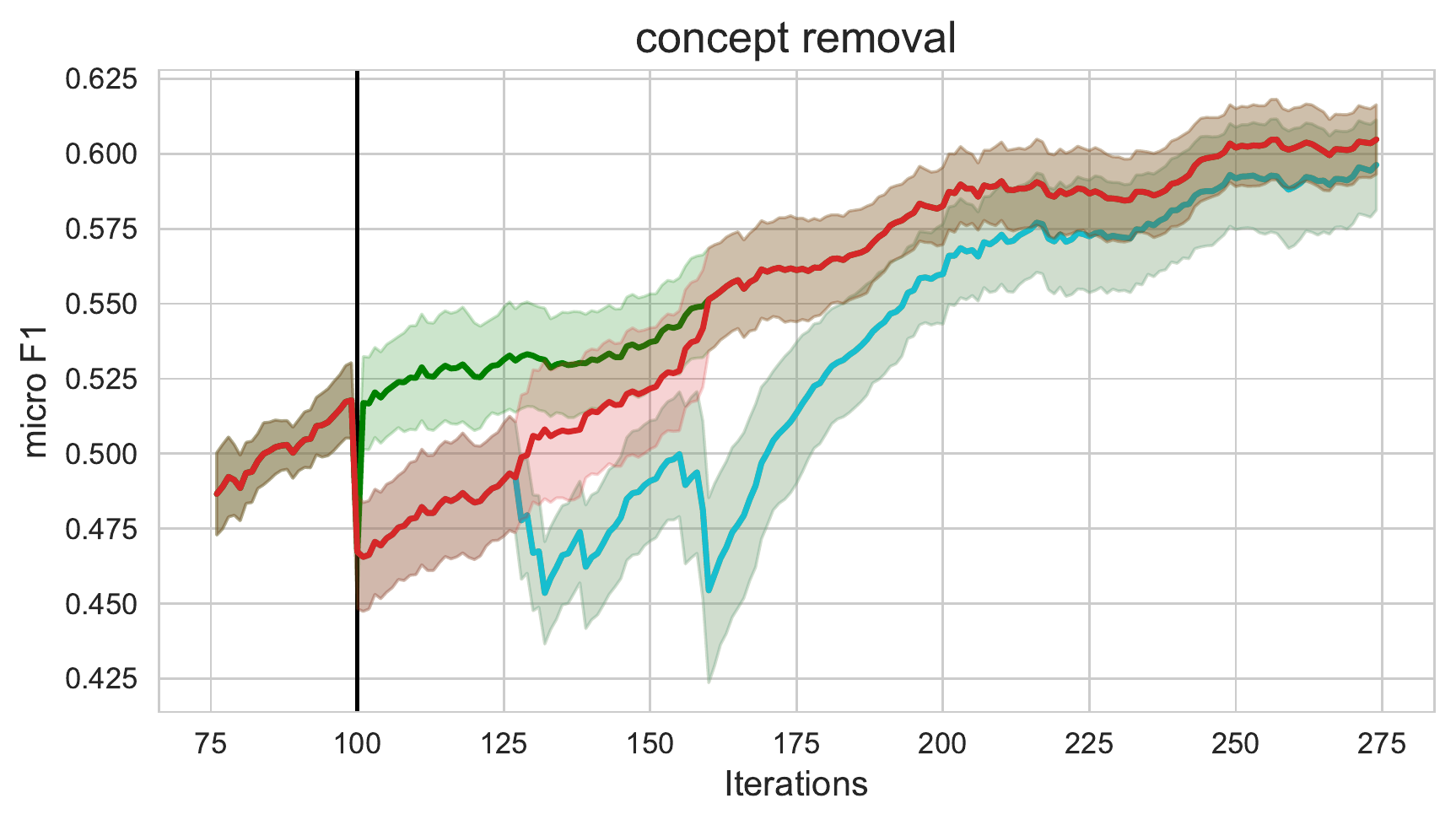}
    \includegraphics[width=0.3\textwidth]{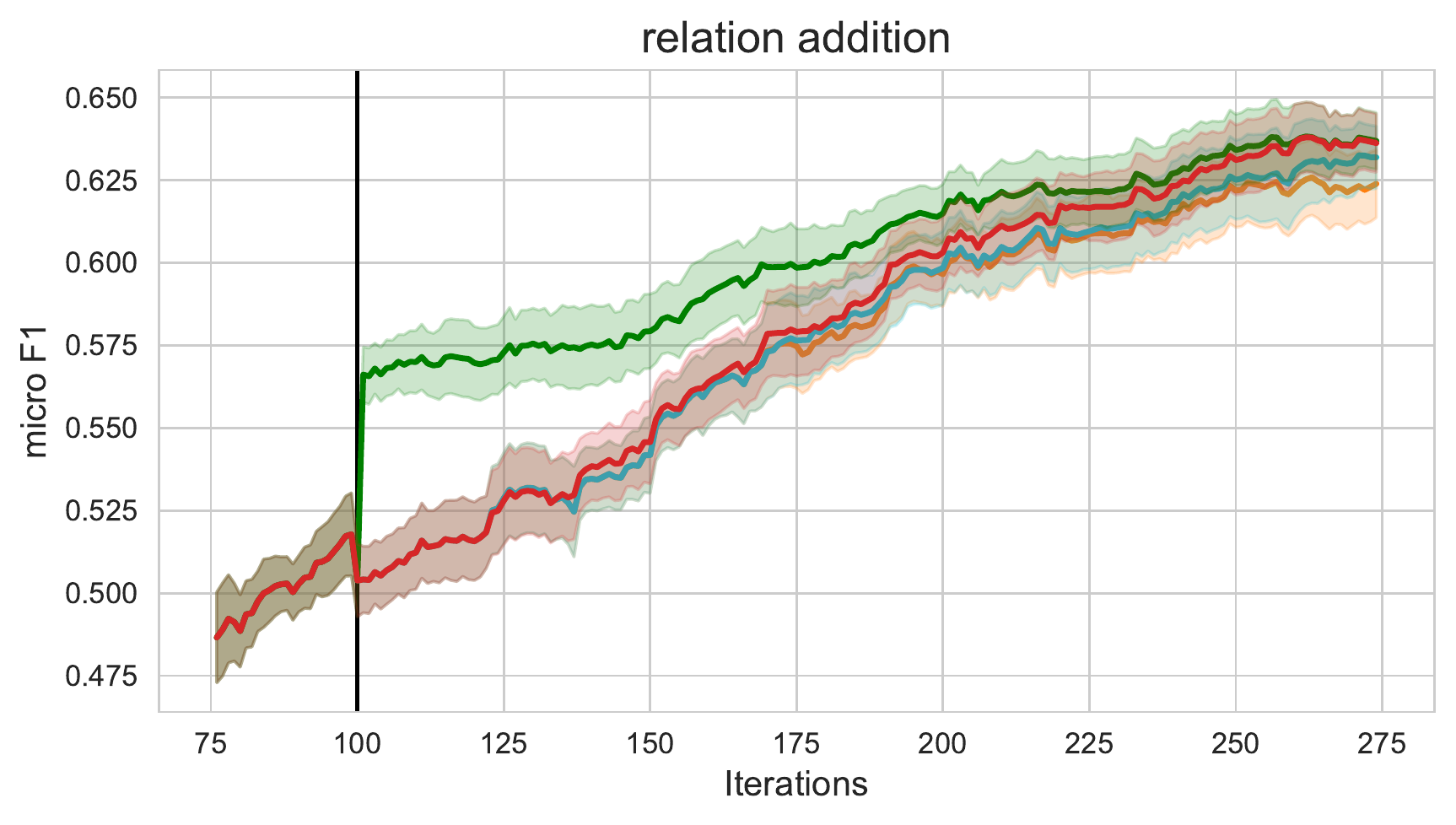}
    \includegraphics[width=0.3\textwidth]{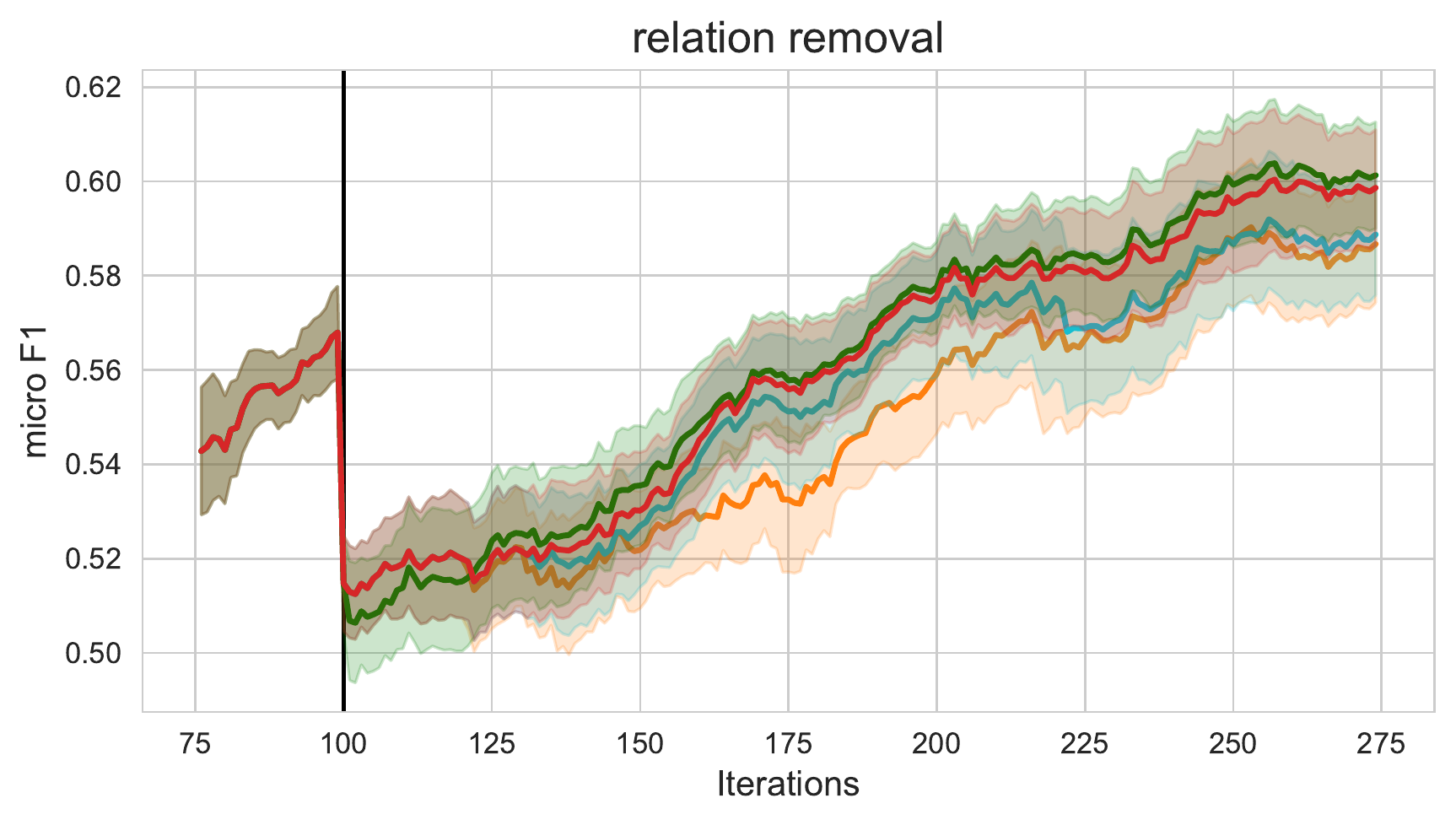}
    
    \caption{Interactive versus fully automated drift disambiguation.  \textbf{Left to right}:  concept removal, relation addition, and relation removal.  \textbf{Top}:  HSTAGGER.  \textbf{Bottom}:  EMNIST.  The results for 20NG can be found in the Supplementary Material due to space constraints.}
    \label{fig:q2}
\end{figure*}

\begin{figure*}[tb]
    \centering
    \begin{tabular}{ccc}
        \includegraphics[width=0.3\textwidth]{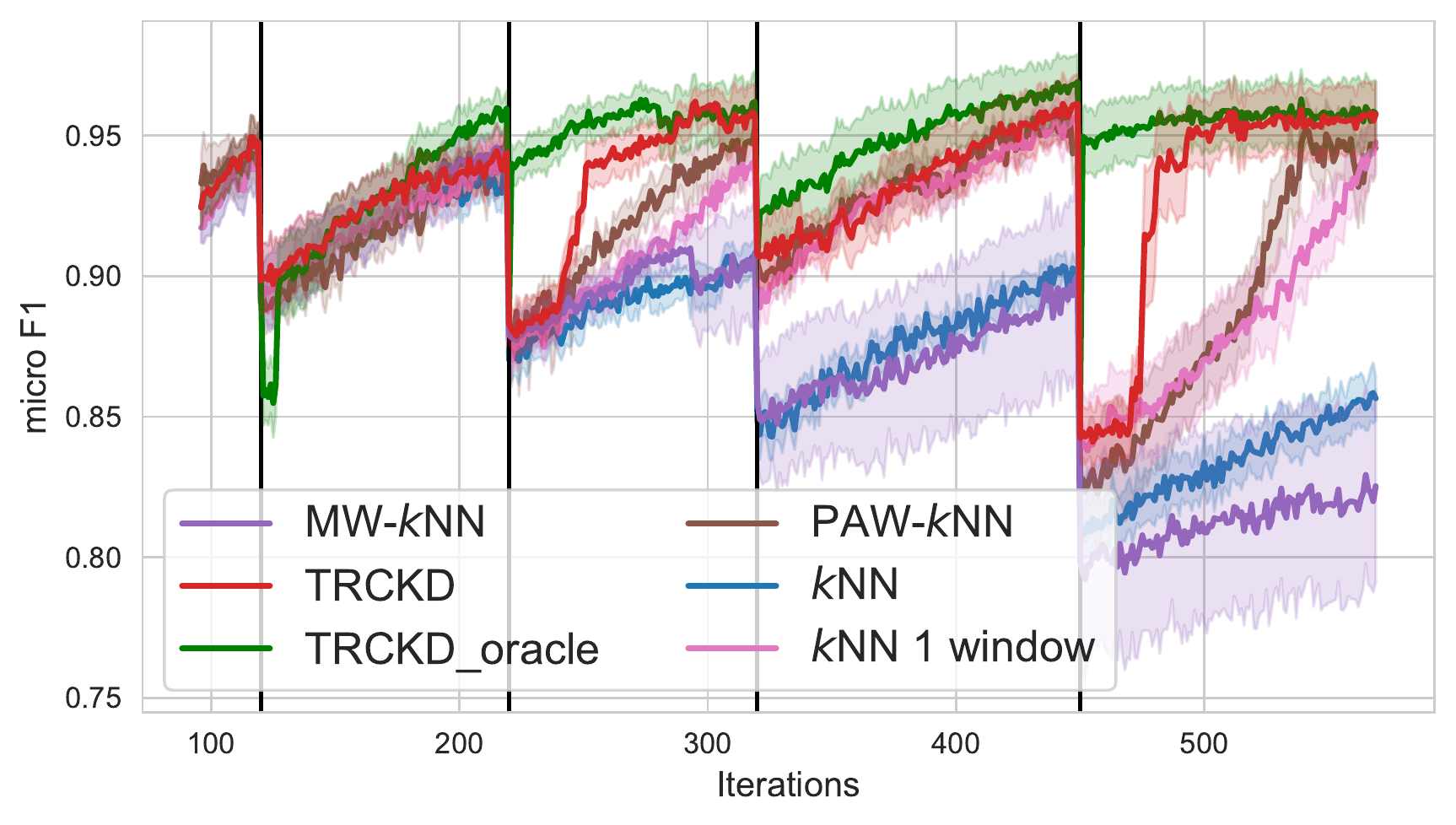} &
        \includegraphics[width=0.3\textwidth]{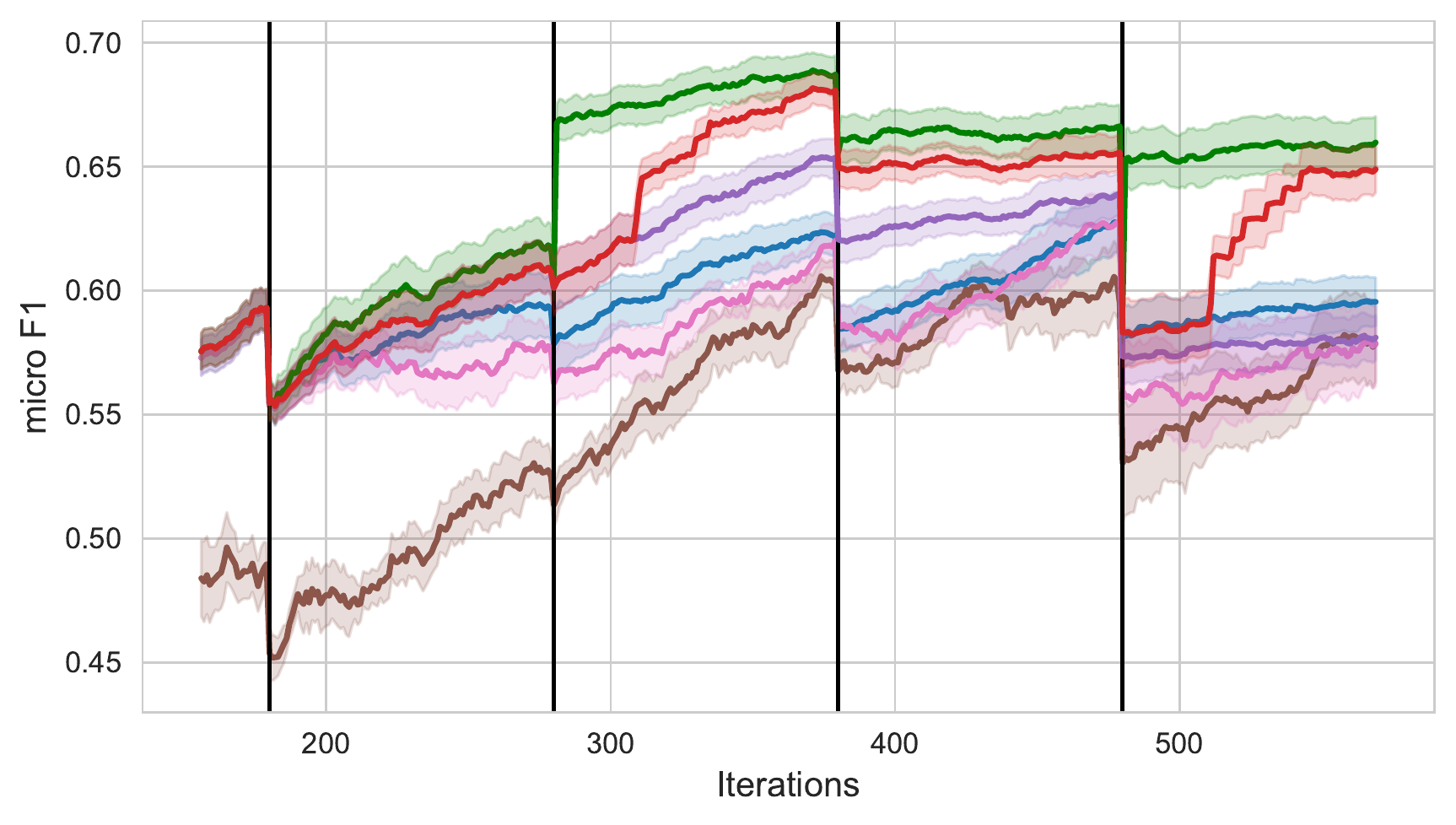} &
        \includegraphics[width=0.3\textwidth]{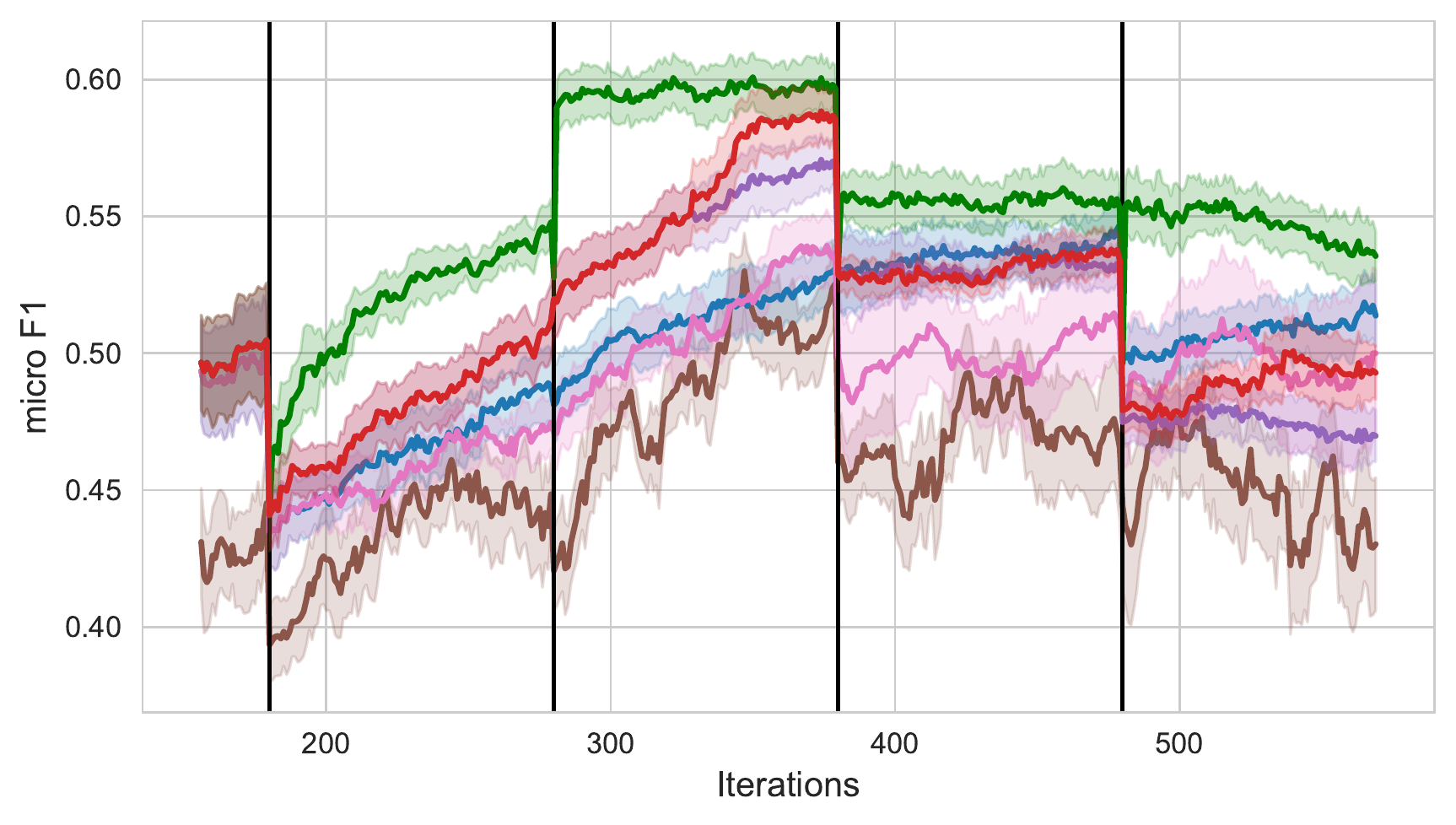}
    \end{tabular}
    
    \caption{\acronym versus competitors on sequential KD.  \textbf{Left to right}:  HSTAGGER, EMNIST, and 20NG.}
    \label{fig:q3}
\end{figure*}

\section{Experiments}

We empirically address the following research questions:
\begin{description}

    \item[Q1] Is knowledge-aware adaptation useful? 

    \item[Q2] Does interacting with the user help adaptation?  
    
    \item[Q3] Does \acronym work in realistic, multi-drift settings?
    
\end{description}
The code for all experiments is at: \url{https://gitlab.com/abonte/handling-knowledge-drift}.

\paragraph{Competitors.}  We compared \acronym against several alternatives:
\textbf{MW-kNN}: the multi-window $k$NN approach of~\cite{spyromitros2011dealing} designed specifically for multi-label problems.
\textbf{PAW-kNN}: punitive adaptive window $k$NN, a state-of-the-art multi-label approach that employs a single window for all concepts and adapts by discarding examples responsible for prediction mistakes~\cite{roseberry2019multi}.
\textbf{$\acronym_{LLR}$}: a fully-automated version of \acronym that follows up MMD detection by performing drift disambiguation with a likelihood ratio test.  This test detects a relation $y^j$ \emph{is-a} $y^i$ iff $\distrib(y^j | y^i) / P(y^j | \neg y^i) \ge \beta$ with\footnote{This is the best possible value for $\beta$ as the ground-truth data is assumed to be always consistent with the ground-truth hierarchy.} $\beta = \infty$.  If the test detects relation addition/removal, $\acronym_{LLR}$ applies the corresponding knowledge-aware adaptation strategy, otherwise it defaults to emptying the current window of the detected concept(s).
\textbf{$k$NN 1-window}: $k$NN with a single window for all concepts that forgets old examples).
\textbf{$k$NN}: regular $k$NN with no adaptation.

\paragraph{Data sets.}  We ran experiments on three data sets.
\textbf{HSTAGGER}: a hierarchical version of STAGGER, a widely used synthetic data set of two-dimensional objects with three categorical attributes (shape, color, size) and labeled by drifting random formulas like ``small and (green or red)''~\cite{schlimmer1986incremental}.  HSTAGGER has $3$ attributes with $4$ values each and labels instances using $5$ different drifting random formulas chosen to have a reasonable pos./neg. ratio. The hierarchy is created by selecting two concepts as part of a third one that acts as parent concept.
\textbf{EMNIST}: a data set of $28 \times 28$ handwritten digits and (uppercase, lowercase) letters~\cite{cohen2017emnist}.  We simulate hierarchical data by running multiple streams of EMNIST examples in parallel.  The digits and letters are grouped in $5$ concepts and structured as in HSTAGGER.  Instances were embedded using a variational autoencoder~\cite{kingma2013auto}.
\textbf{20NG}: a data set of newsgroup posts categorized in twenty topics.\footnote{From archive.ics.uci.edu/ml/datasets/Twenty+Newsgroups}  The data set was converted to hierarchical classification by grouping different classes into super-topics (e.g., alt.atheism, soc.religion.christian and talk.religion.misc were grouped into a ``religion'' super-topic).  The documents were embedded using a pre-trained Sentence-BERT model~\cite{reimers2019sentence} and compressed to $100$ features using PCA.
The data sets are converted into streams by sequentially sampling random examples.

\paragraph{Experimental details}  All experiments were run on a machine with eight 2.8 GHz CPUs and 32 GiB RAM.  Each experiment was run ten times by changing the random seed:  2 for selecting hyperparameters and 8 for evaluating performance.
Our evaluation focuses on concept deletion, relation addition, and relation removal;  plain concept drift is reported in the Supplementary Material. %
In the three cases, KD is injected into the stream by removing a random concept from the available ones, adding a random relation, and removing a random relation, respectively.  KD starts after the competitors approximately reach their peak performance, namely after $100$ iterations for HSTAGGER and EMNIST, and $170$ for 20NG.
Performance was measured in terms of micro $F_1$ score on a hold-out test set (of size $64$ for HSTAGGER and $200$ for EMNIST and 20NG) randomly selected before each run.  All methods observed the same sequence of examples and drifts.  The plots report standard error information.  User replies were simulated by an oracle that always answers correctly to disambiguation queries.

\paragraph{Hyperparameters}  For \acronym, the window size $w$ was fixed to 200 so to make it perform as well as standard $k$NN when no drift is present.  MW-$k$NN and all variants of \acronym use the same window size.
To speed up detection, the MMD is computed on the 70 most recent examples only.  Prior to each experiment, the threshold $\tau$ was selected from $\{0.4, 0.5\}$ to optimize $F_1$ on two independent runs.
The ranges of $k$ differ between methods, the detailed values are reported in the Supplementary Material.
For PAW-$k$NN, the penalty ratio was set to $p=1$ as suggested by~\citet{roseberry2019multi} and the minimum and maximum window sizes to $50$ and $200$ for consistency with \acronym.

\paragraph{Q1:  Is knowledge-aware adaptation useful?}  To evaluate the impact of our adaptation strategy, we remove unrelated effects due to spurious/delayed detection by telling all approaches exactly when KD occurs.  We compare \textbf{$\acronym_{oracle}$} (i.e., \acronym that knows exactly when drift occurs) to MW-$k$NN, PAW-$k$NN, $k$NN 1-window, regular $k$NN and \textbf{$\acronym_{forget}$} (\acronym that adapts to all types of KD by forgetting old examples).

The results can be viewed in~Figure~\ref{fig:q1}.  The plots show that $\acronym_{oracle}$ is by far the best performing method on all data sets and for all forms of KD.  In most cases the runner up is $\acronym_{forget}$:  while it performs similarly to $\acronym_{oracle}$ for concept drift and removal (see the Supplementary Material;  this is because our adaptation strategy boils down to forgetting in this simple setup), it does lag behind for relation addition and removal, showing a sizeable advantage for knowledge-aware adaptation.  MW-$k$NN works reasonably well but suffers from relying on passive adaptation and does not always performs better than the two $k$NN baselines.  PAW-$k$NN tends to underperform on EMNIST and 20NG, especially when KD affects the relations.  These results validate knowledge-aware adaptation on all data sets and allows us to answer \textbf{Q1} in the affirmative.  For this reason, we will focus on knowledge aware adaptation in the following experiments.

\paragraph{Q2:  Is interaction useful?}  To measure the impact of interaction, we compare four variants of \acronym that differ in what information they elicit from the supervisor, namely: \acronym, $\acronym_{LLR}$,
$\acronym_{oracle}$
and \textbf{$\acronym_{ni}$} (like \acronym except that instead of interacting with the user it assumes that all concepts detected as drifting by MMD have undergone individual concept drift and adapts by purging their current window).
The results in Figure~\ref{fig:q2} are quite intuitive:  \textbf{$\acronym_{oracle}$} substantially outperforms all alternatives in all cases except for relation removal in 20NG.  This shows that, if drift is detected correctly and timely, interactive disambiguation is extremely useful for guiding knowledge-aware adaptation and quickly aligning the model to the ground-truth.  \acronym tends to perform substantially better than the no-interaction baselines $\acronym_{LLR}$ and $\acronym_{ni}$ and if MMD detection works well it quickly reaches the performance of the oracle.  This allows us to answer \textbf{Q2} in the affirmative.   If MMD underperforms (as in EMNIST), \acronym does not get a chance to quickly interact with the user and shows no improvement.  This could be fixed by better optimizing the choice of kernel and threshold used by MMD, perhaps by turning them into per-concept parameters.  This is left to future work.  Importantly, \acronym interacts with the user $1.54 \pm 0.78$ times per run on average, showing that few interaction rounds are often enough to achieve a noticeable performance boost.

\paragraph{Q3: Does \acronym work in multi-drift settings?}  We consider a realistic scenario with four sequential KD events, namely concept drift, relation addition, relation removal, and concept removal.  The results in Figure~\ref{fig:q3} show that \acronym tends to outperform all competitors except the oracle.  The advantage is quite marked whenever the KD affects the concept hierarchy itself, up to $+10\%$ $F_1$ for HSTAGGER and $+5\%$ for EMNIST.  The plots mirror the advantages shown by \acronym in the previous experiments and highlight that the benefits knowledge-aware adaptation and interaction carry over to more realistic settings.  This allows us to answer \textbf{Q3} in the affirmative.  The lack of reactive adaptation penalizes MW-$k$NN and PAW-$k$NN, the latter especially on EMNIST.

\section{Related Work}

\acronym is the first approach that considers KD in hierarchical classification and that explicitly tackles drift disambiguation.  Indeed, existing work on concept drift focuses on the single-label~\cite{gama2014survey} and multi-label~\cite{zheng2019survey} cases, while work on drift understanding is unconcerned with hierarchical classification~\cite{lu2018learning}.  \acronym is also the first approach to tackle drift disambiguation by interacting with a human supervisor.
Sliding windows were first employed for drift detection in~\cite{kifer2004detecting}.  This setup offers distribution-free guarantees on detection accuracy under mild assumptions.  The idea of using examples in the windows for explanatory purposes was also discussed in~\cite{kifer2004detecting}.  The MMD, which \acronym uses to compare present and past data, was leveraged in domain adaptation~\cite{zhang2013domain}.  More recent two-sample tests, like ME~\cite{jitkrittum2016interpretable}, underperformed in our experiments.

Open world recognition is a streaming classification setting in which unanticipated classes appear over time~\cite{boult2019learning,bendale2015towards}. Open world recognition only deals with concept addition, and it is unconcerned with more general KD.
Other recent work on interactive classification under concept addition~\cite{bontempelli2020learning} focuses on handling noisy labels rather than on adapting to drift.

Finally, in continual learning the machine must learn new concepts (or tasks) over time.  The key difference is that our goal is to smartly forget obsolete information, whereas in continual learning the main issue is how to prevent the machine from forgetting previously acquired knowledge~\cite{parisi2019continual}.  More generally, continual learning neglects other forms of KD altogether.

\section{Conclusion}

We introduced the problem of knowledge drift in hierarchical classification and proposed to partially offload drift disambiguation to a user.  We introduced \acronym, an approach for learning under KD that combines \emph{automated} drift detection and adaptation, upgraded to hierarchical classifiers, with \emph{interactive} drift disambiguation.   Our results indicate that \acronym outperforms fully automated approaches by asking just a few questions, even when detection performance is not ideal.

In future work, we plan to improve the interpretability of our interaction protocol by integrating ideas from explainable AI~\cite{demvsar2018detecting,schramowski2020making},
generalize \acronym to active learning settings in which the labels must be explicitly queried
and  develop KD adaptation strategies for neural nets in continual learning.

\section{Acknowledgments}

The research of FG and AP has received funding from the European Union's Horizon 2020 FET Proactive project ``WeNet -- The Internet of us'', grant agreement No 823783.  The research of AB and ST has received funding from the ``DELPhi - DiscovEring Life Patterns'' project funded by the MIUR Progetti di Ricerca di Rilevante Interesse Nazionale (PRIN) 2017 -- DD n. 1062 del 31.05.2019.

\bibliography{paper}

\subsection*{Additional Details}

\paragraph{Hyperparameters.}  Table~\ref{tab:hyperparameters} reports the values used in each experiment.

\begin{table}[h]
    \centering
    \begin{tabular}{lllll}
        \toprule
        \textbf{RQ} & \textbf{Drift} & $\tau$ & \textbf{\ACRONYM} $k$ & \textbf{PAW-$k$NN} $k$
        \\
        \midrule
        \multicolumn{5}{c}{\textbf{HSTAGGER}}
        \\
        \midrule
        \multirow{4}{*}{Q1 - Q2} & CD    & 0.04  & 3     & 3     \\ \cmidrule{2-5}
                            & CR    & 0.04  & 11    & 3     \\ \cmidrule{2-5}
                            & RA    & 0.04  & 11    & 3     \\ \cmidrule{2-5}
                            & RR    & 0.04  & 11    & 3     \\ \midrule
        Q3                  & All   & 0.04  & 11     & 3
        \\
        \midrule
        \multicolumn{5}{c}{\textbf{EMNIST}}
        \\
        \midrule
        \multirow{4}{*}{Q1 - Q2} & CD    & 0.04  & 5     & 3     \\ \cmidrule{2-5} 
                            & CR    & 0.05  & 3     & 3     \\ \cmidrule{2-5} 
                            & RA    & 0.05  & 3     & 3     \\ \cmidrule{2-5} 
                            & RR    & 0.04  & 3     & 3     \\ \midrule
        Q3                  & All   & 0.05  & 3     & 3
        \\
        \midrule
        \multicolumn{5}{c}{\textbf{20NG}}
        \\
        \midrule
        \multirow{4}{*}{Q1 - Q2} & CD    & 0.05  & 3     & 3     \\ \cmidrule{2-5} 
                            & CR    & 0.05  & 3     & 3     \\ \cmidrule{2-5} 
                            & RA    & 0.04  & 3     & 3     \\ \cmidrule{2-5} 
                            & RR    & 0.04  & 3     & 3     \\ \midrule
        Q3                  & All   & 0.05  & 3     & 3
        \\
        \bottomrule
    \end{tabular}
    \caption{\label{tab:hyperparameters}  Hyperparameter values.}
\end{table}

\noindent The hyperparameters are:  MMD threshold $\tau$ used by \acronym;  number of neighbors $k$ used by \acronym and all competitors except PAW-$k$NN;  and number of neighbors $k$ used by PAW-$k$NN.  Abbreviations:  CD is concept drift, CR concept removal, RA relation addition, and RR relation removal.
Before each experiment, the value of $\tau$ and $k$ are selected in two independent runs by optimizing $F_1$. The value of $\tau$ is selected from $\{0.4, 0.5\}$;  these values were frequently observed to indicate drift in our experiments.  The value of $k$ from $\{3, 5, 11\}$ and it was chosen independently for \acronym and its variants, including MW-$k$NN, and for PAW-$k$NN.

All other hyperparameters remain fixed across experiments.
The window size of \acronym was set to 200.
The penalty ratio of PAW-$k$NN was set to $p=1$ as in~\cite{roseberry2019multi}, while the minimum and maximum window sizes were set to $m_{min}=50$ for HSTAGGER and EMNIST and 80 for 20NG data set and $m_{max}=200$ repsectively.
To speed up detection, the MMD is only computed on the 70 most recent examples.  This choice does not sacrifice reliability of detection.

The inference and training time is similar for all methods since they are all based on $k$NN. One method requires ~2 seconds for training on the new example and evaluating on test set at each iteration.

\paragraph{Data sets.}  Table~\ref{tab:datasets} reports, for each of our three data sets, the number of examples sampled to generate the stream $|S|$, the number of attributes $d$ and their type, the number of concepts $c$, as well as the following three measures of annotation density:  the (average) number of positive labels per example $LC$, the empirical probability that a label is positive $LD$, and how many distinct combinations of positive categories (out of $2^c$) are annotated in the data $DL$.  These metrics are taken from~\cite{zhang2010multi}.  All instances in the data set belong to the \textit{root} concept.

\begin{table*}[tb]
    \centering
    \begin{tabular}{lccccccc}
        \toprule
        \textbf{Data set}   & $|S|$ & $d$   & \textbf{Type} & $c$   & $LC$              & $LD$              & $DL$ \\
        \midrule
        HSTAGGER        & 570   & 3     & cat.          & 6     & 4.42 $\pm$ 0.16   & 0.63 $\pm$ 0.02   & 34.63 $\pm$ 11.89 \\
        EMNIST              & 570   & 10    & cont.         & 9     & 3.5 $\pm$ 0.04   & 0.44 $\pm$ 0.00   & 55.38 $\pm$ 5.57  \\
        20NG                & 570   & 100   & cont.         & 6     & 2.19 $\pm$ 0.00   & 0.31 $\pm$ 0.00   & 8.5 $\pm$ 0.5 \\
        \bottomrule
        \end{tabular}
        \caption{Data sets statistics (mean $\pm$ std.). $|S|$: number of instances, $d$: number of attributes, type of features (categorical or continuous), $c$: number of labels, $LC$: label cardinality, $LD$: label density, $DL$: distinct label set.  The metrics are averaged on 8 runs and refer to the experiment for Q3.}
\label{tab:datasets}
\end{table*}

\paragraph{Metrics.}  The plots in the paper focus on the micro $F_1$ score, which consider the sparsity of the classes of our scenario.
Letting $\{(\vx_i, \vy_i) : i = 1, \ldots, n\}$ be the examples in the \emph{test set}, $\hat{\vy}_i$ their predictions and $y_i^j$ the $j$th element of $y_i$, which is 1 if $\vx_i$ belong to the $j$th concept and 0 otherwise, the micro $F_1$ is defined as follows:
\begin{align*}
    F_{\text{$1$-micro}}
        &= \frac{2 \sum_{j=0}^c \sum_{i=1}^n y_i^j \hat{y}_i^j}{\sum_{j=0}^c \sum_{i=1}^n y_i^j + \sum_{j=0}^c \sum_{i=1}^n \hat{y}_i^j}
\end{align*}
The results showed in the following figures are averaged on 8 runs and for each of them the standard error is reported.

\subsection*{MMD Versus Mean Embeddings}

In order to evaluate whether MMD is fit for dealing with KD, we implemented Mean Embeddings (ME), an state-of-the-art kernel-based discrepancy between distributions~\cite{jitkrittum2016interpretable}.  Importantly, the discrimination power of ME can be maximized on an independent training set with gradient ascent.  In our experiment, we carried out this optimization on the pre-drift examples (i.e., on the first 100 examples in the stream for HSTAGGER and EMNIST and on the first 170 for 20NG).  We also used the very same tensor product kernel for both MMD and ME and for ME we tuned the width of the Gaussian kernel $k_X$ over instances along with the discrimination power of ME.

\begin{figure}
     \centering
     \includegraphics[width=0.33\textwidth]{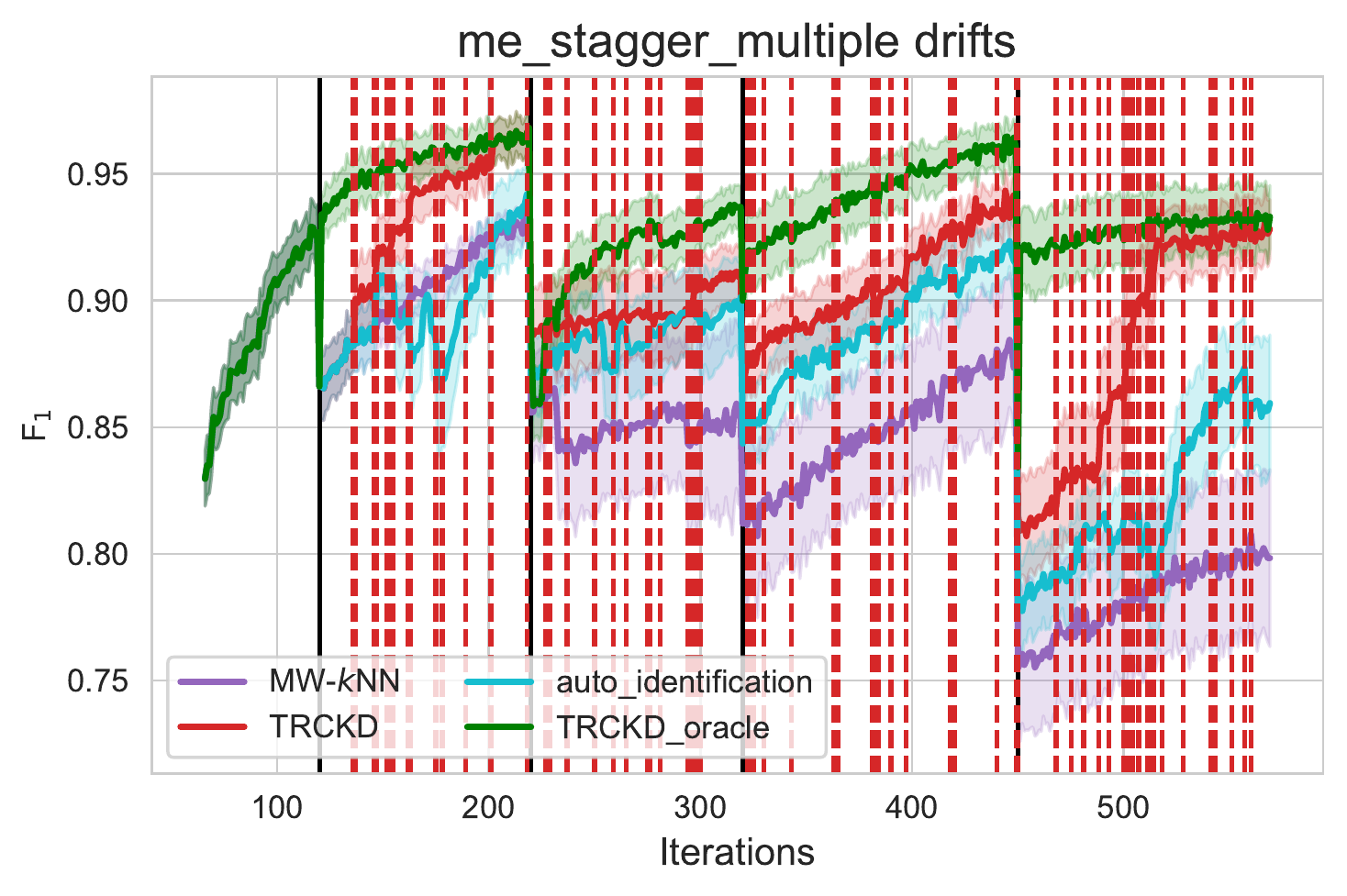}
     \caption{Results for multiple drifts scenario for HSTAGGER. \acronym implements the ME test for drift detection. The ME hyperparameters are $J=5$ and $\alpha=0.01$.}
     \label{fig:metest}
\end{figure}

Despite being more powerful than MMD on paper, the ME did not perform as well in practice.  In particular, ME turned out to be overly sensitive and had severe false detection issues.  The changepoints detected by \acronym+ME are illustrated as vertical dashed bars in~Figure~\ref{fig:metest} for the case of HSTAGGER with sequential KD.  In short, ME tends to detect three or four times as many drifts as MMD.  This makes it inadequate for interacting with the user:  indeed, querying the user too frequently is likely to rapidly make her lose interest in the interaction in practice.

\subsection*{Comparison to Graphical Lasso}

Given the similarity between drift disambiguation and structure learning for probabilistic graphical models~\cite{koller2009probabilistic}, we evaluated a variant of \acronym that uses graphical lasso to reconstruct the structure of the hierarchy from the most recent examples~\cite{friedman2008sparse}.  In particular, in each iteration a data set is built by combining the 70 most recent examples (analogously to what is done for MMD) for all concepts in the machine's hierarchy.  This data set set is fed to graphical lasso, which spits out an (sparse) undirected graph based on the empirical correlation between all the concepts.  The directions of individual edges are set so to maximize the likelihood of the child implying the parent and edges are treated as \emph{is-a} relations.  The resulting directed graph replaces the machine's concept hierarchy.  The difference between the previous and current concept hierarchy is used to perform knowledge-aware adaptation.

\begin{figure}[tb]
    \centering
    \includegraphics[width=0.30\textwidth]{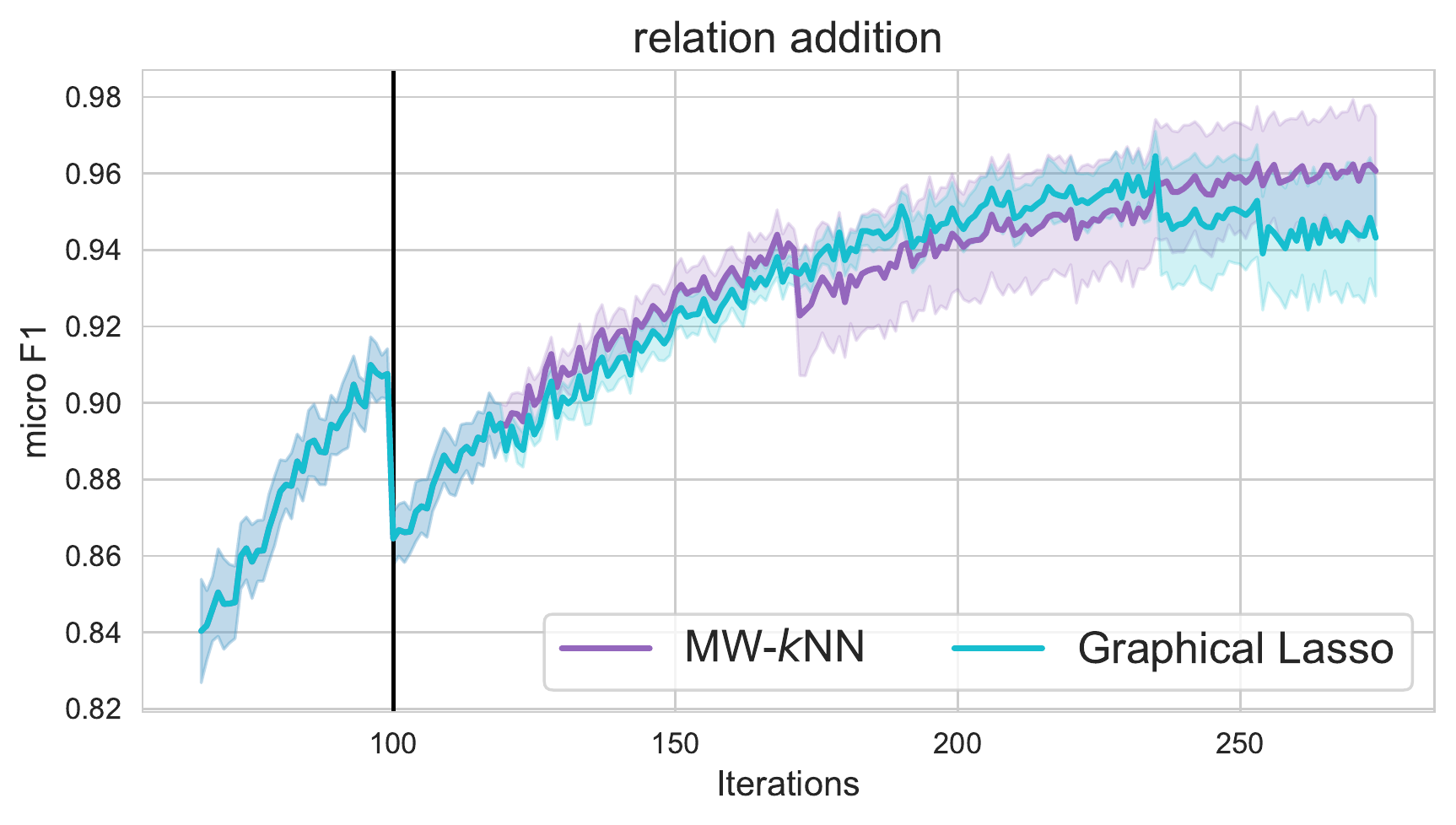}
    \includegraphics[width=0.30\textwidth]{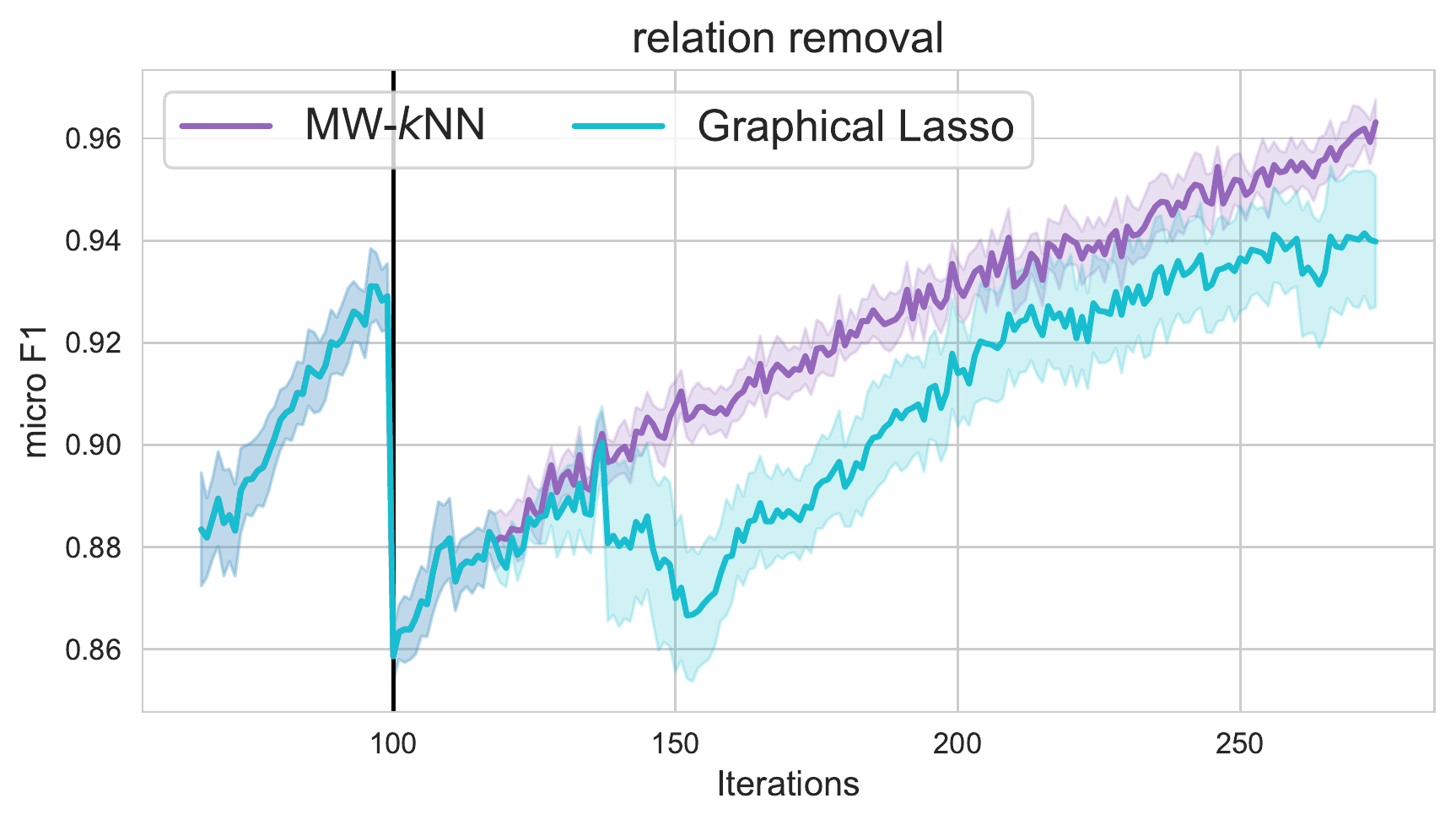}
    \caption{HSTAGGER, automatic drift type identification with Graphical Lasso and MMD for detection. Top: relation addition. Bottom: relation removal.}
    \label{fig:auto_lasso_hstagger}
\end{figure}

A comparison between \acronym+lasso and a baseline MW-$k$NN is reported in Figure~\ref{fig:auto_lasso_hstagger}.  It turns out that for relation addition and removal, graphical lasso often fails to estimate the ground-truth concept hierarchy, leading to systematic prediction errors.  Furthermore, it is quite unstable and often detects spurious changes to the hierarchy.  The main issue is that -- like other fully automated approaches for structure learning -- graphical lasso does require substantial amounts of data to perform reliably, and this is simply not the case in our non-stationary setting.  This makes structured learning-based approaches unsuitable for this setup.

\subsection*{Full Plots for Q1}

\begin{figure*}[tb]
    \centering
    \begin{tabular}{cccc}
        \includegraphics[width=0.23\textwidth]{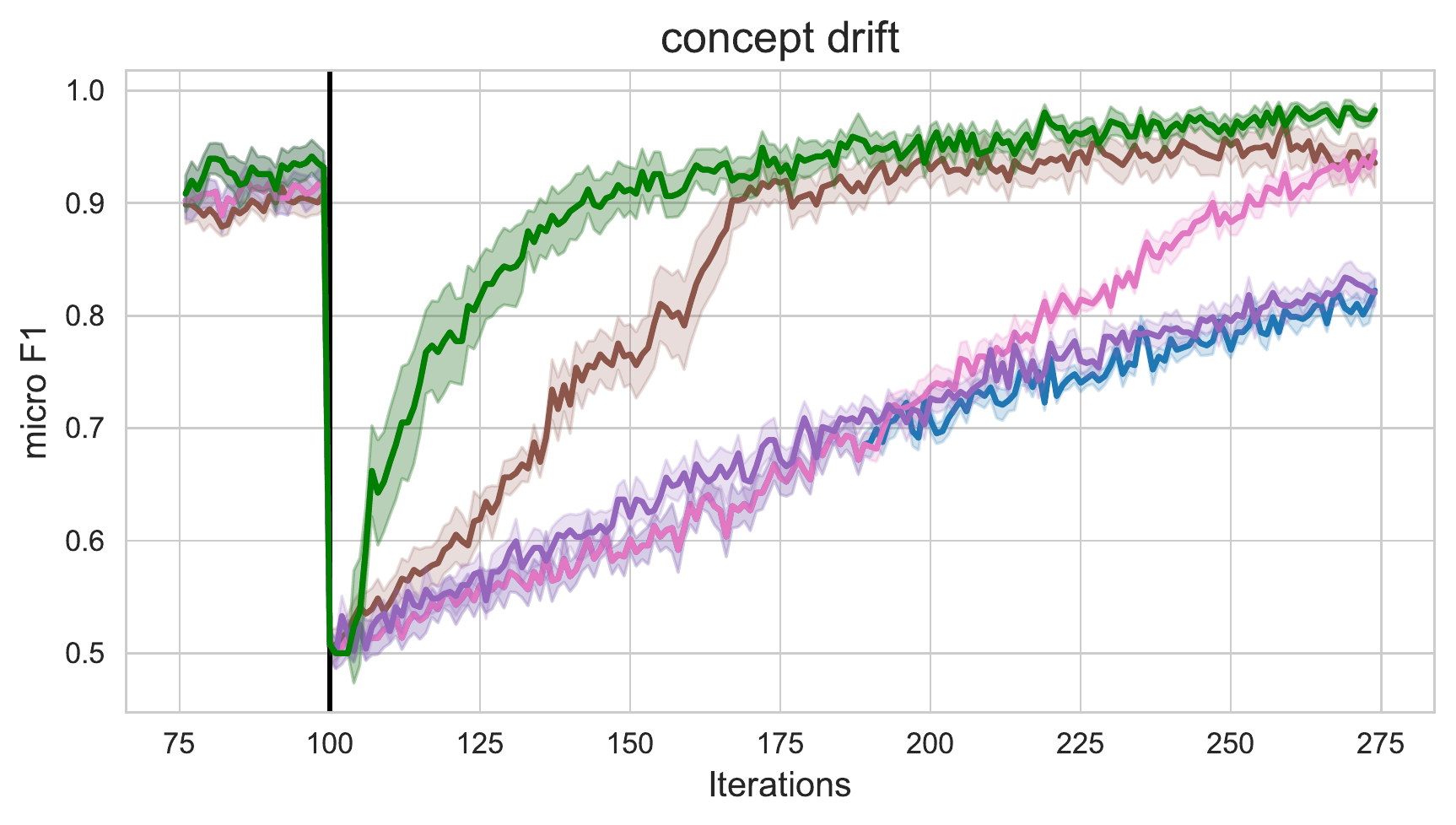} &
        \includegraphics[width=0.23\textwidth]{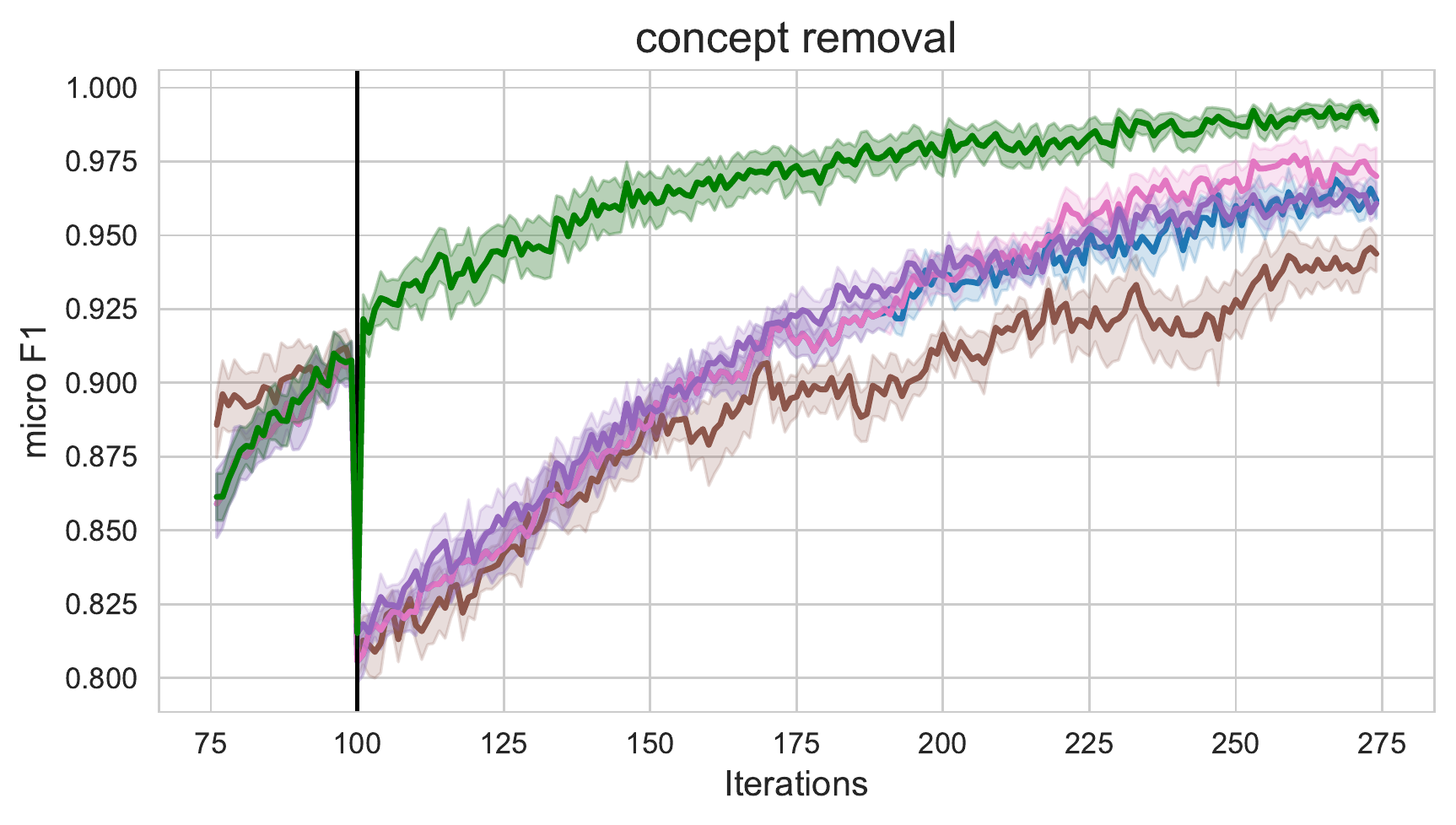} &
        \includegraphics[width=0.23\textwidth]{figures/rq1/f1micro_stagger_add_implication.pdf} &
        \includegraphics[width=0.23\textwidth]{figures/rq1/f1micro_stagger_remove_implication.pdf} \\
        
        \includegraphics[width=0.23\textwidth]{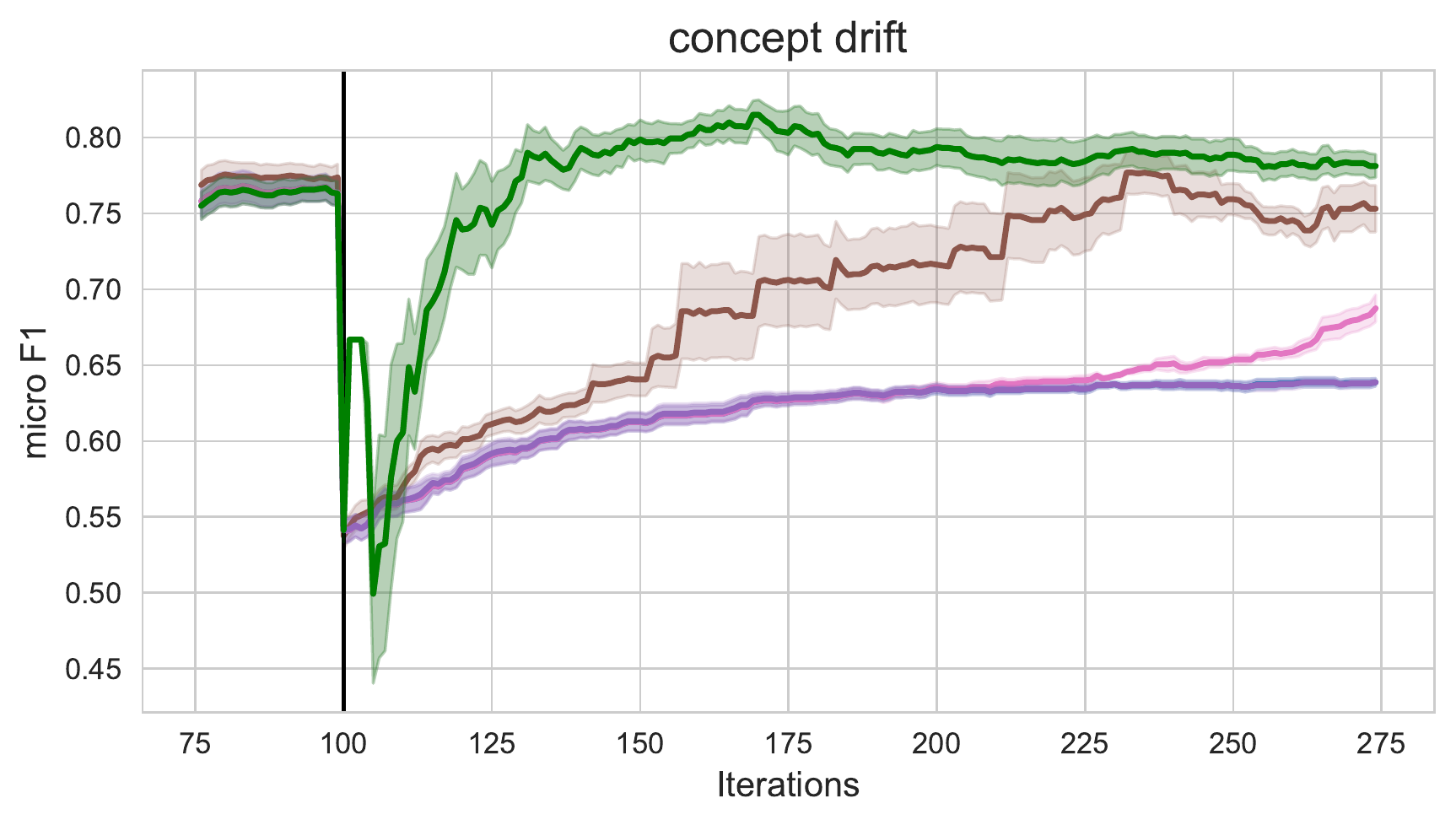} &
        \includegraphics[width=0.23\textwidth]{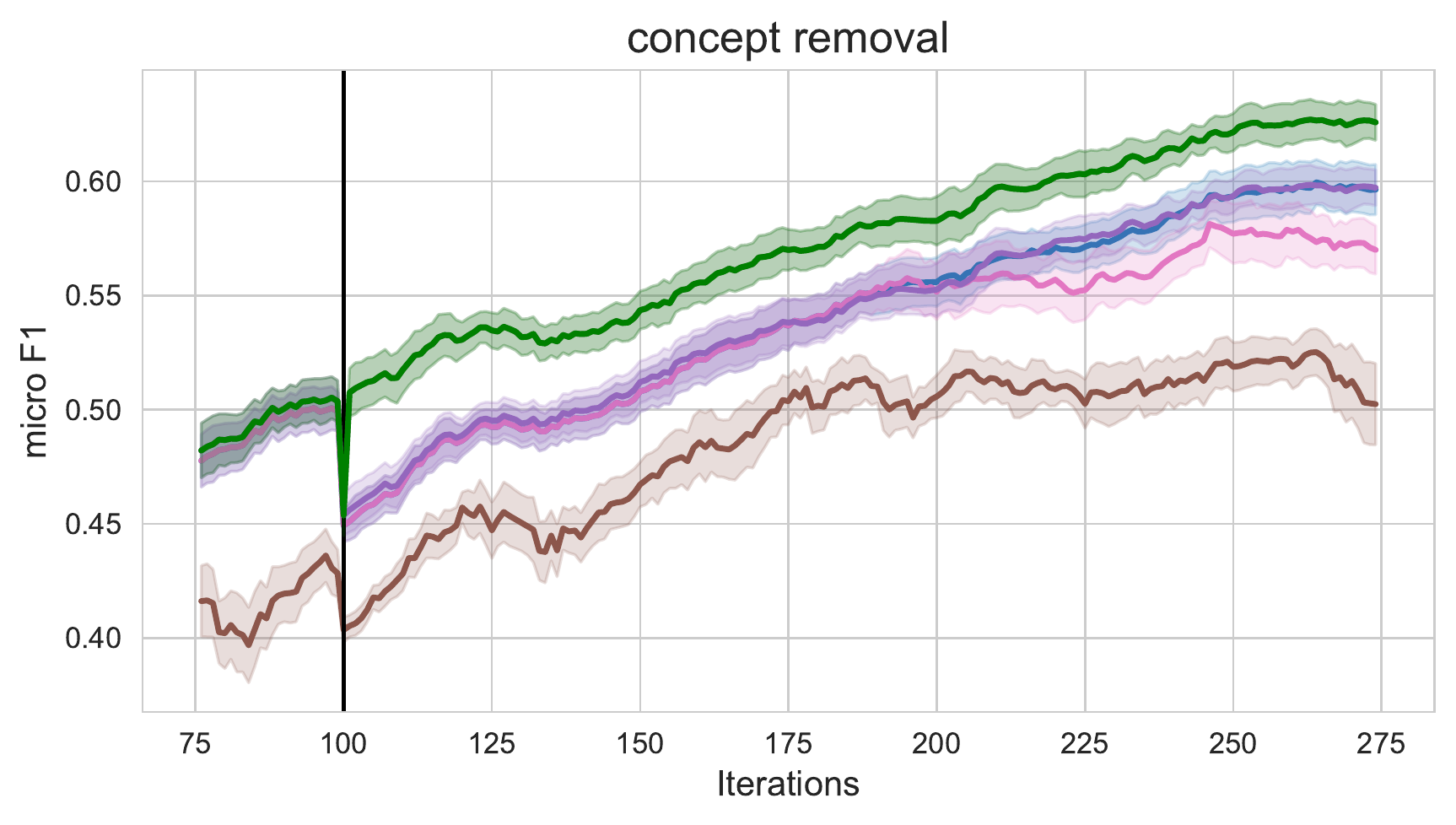} &
        \includegraphics[width=0.23\textwidth]{figures/rq1/f1micro_emnist_add_implication.pdf} &
        \includegraphics[width=0.23\textwidth]{figures/rq1/f1micro_emnist_remove_implication.pdf} \\
        
        \includegraphics[width=0.23\textwidth]{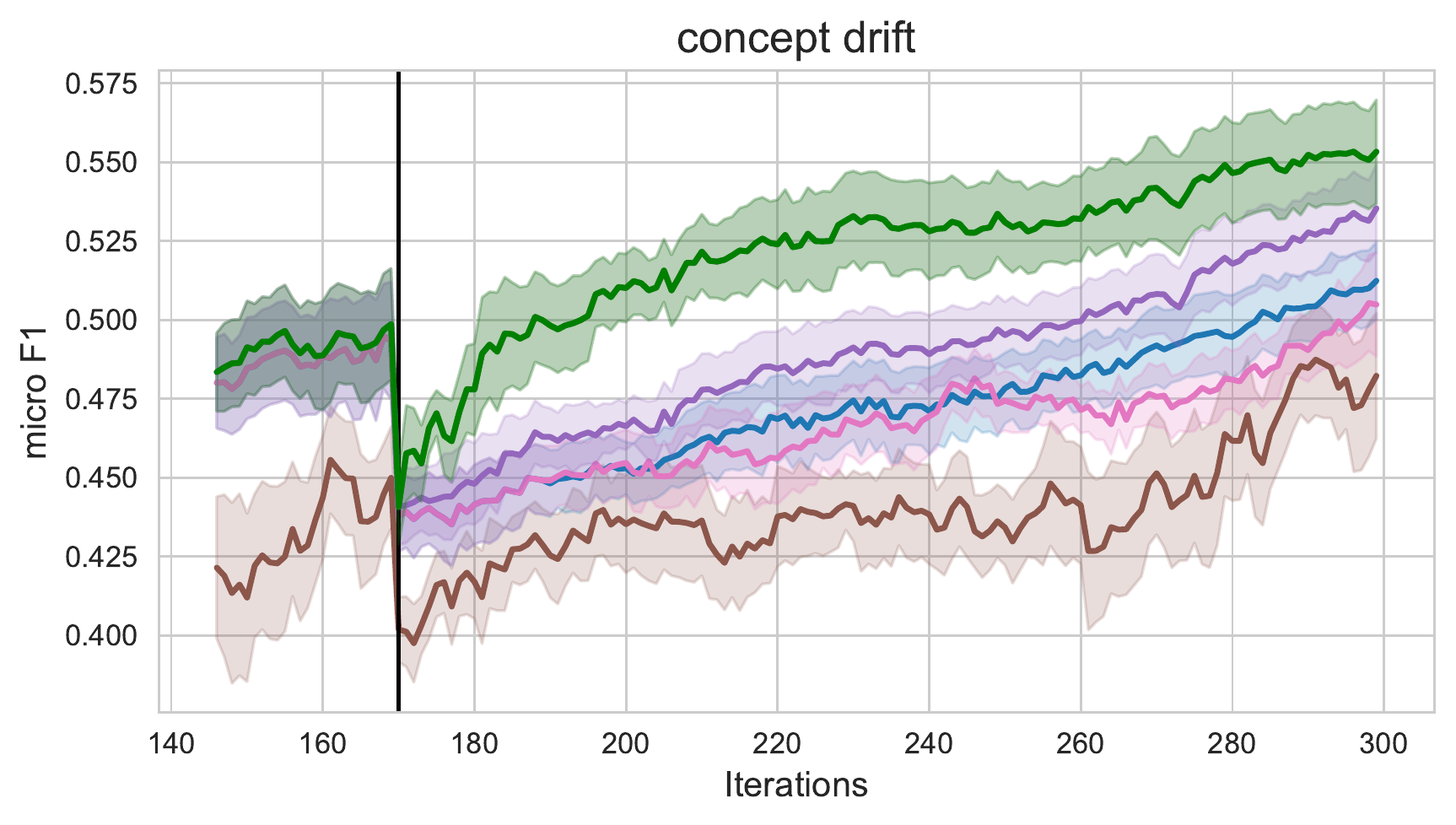} &
        \includegraphics[width=0.23\textwidth]{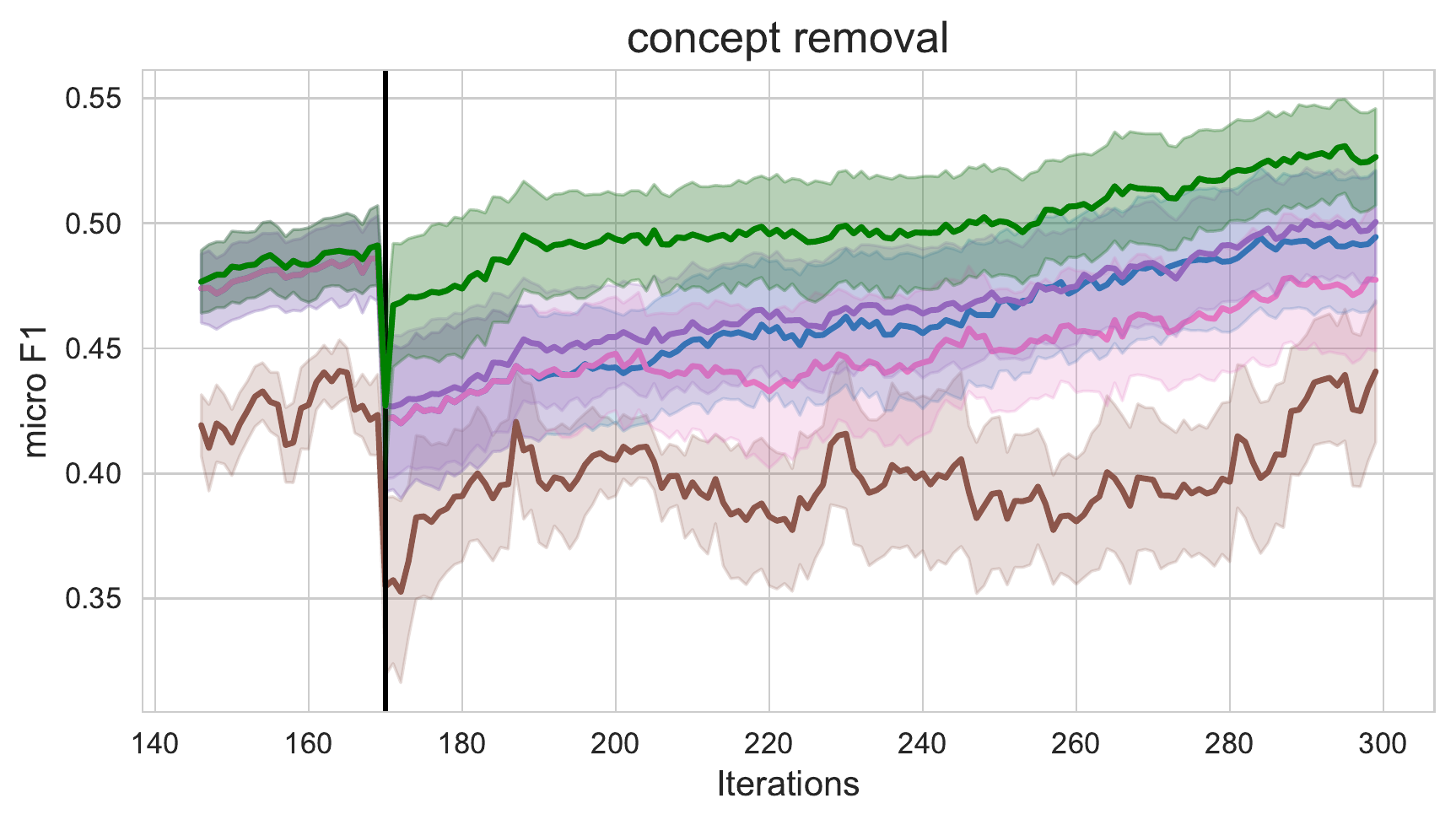} &
        \includegraphics[width=0.23\textwidth]{figures/rq1/f1micro_news_add_implication.pdf} &
        \includegraphics[width=0.23\textwidth]{figures/rq1/f1micro_news_remove_implication.pdf} \\
    \end{tabular}
    \caption{Comparison in terms of micro F$_1$ between \acronym and standard forgetting strategies for $k$NN-based methods.  Top to bottom:  results for HSTAGGER, EMNIST, and 20NG.  Left to right:  concept drift, concept removal, relation addition, and relation removal.}
    \label{fig:q1f1}
\end{figure*}

Figures~\ref{fig:q1f1} report the $F_1$ score for all competitors on all data sets.  The results are the same as in the main text:  $\acronym_{oracle}$ is by far the best performing method and the runner up is $\acronym_{forget}$, which lags behind when dealing with relation addition and removal.  The MW-$k$NN baseline performs well but suffers from relying on passive adaptation.  PAW-$k$NN tends to perform well for concept drift on HSTAGGER and EMNIST, but falls behind on 20NG and for all other forms of drift.

\subsection*{Full Plots for Q2}

To measure the impact of interaction, we compare four variants of \acronym that differ in what information they elicit from the supervisor:
$\acronym_{oracle}$ which knows exactly when and what kind of knowledge drift occurred;
\acronym, our proposed approach that combines MMD for detection and interactive identification;
$\acronym_{ni}$ (no interaction) that trusts MMD for both detection and disambiguation, and uses forgetting for adaptation since MMD cannot distinguish between concept drift and relation addition/removal;
$\acronym_{LLR}$: a fully-automated version of \acronym that follows up MMD detection by performing drift disambiguation with a likelihood ratio test.

\begin{figure*}[tb]
    \centering
    \begin{tabular}{cccc}
        \includegraphics[width=0.24\textwidth]{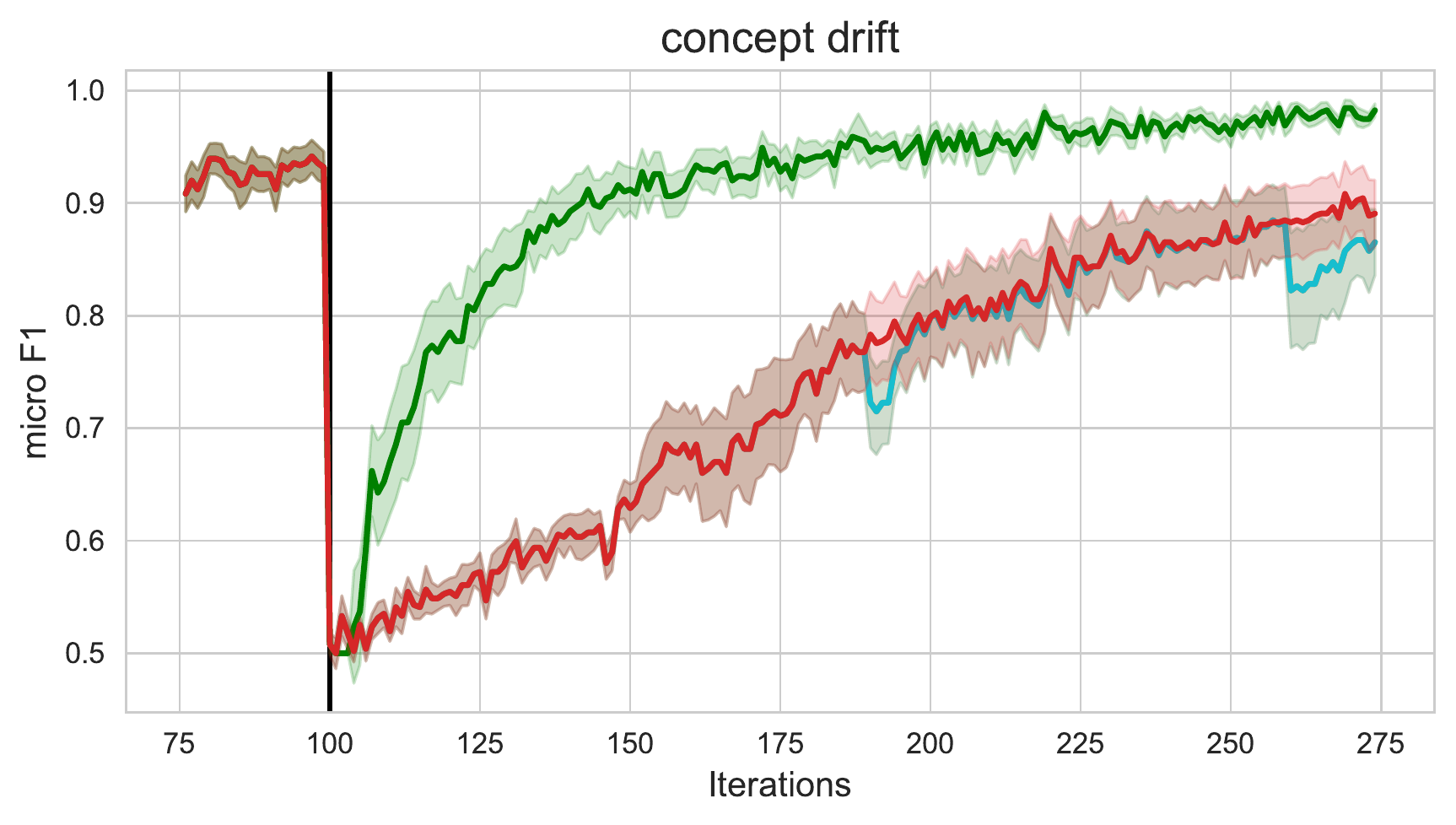}
        \includegraphics[width=0.24\textwidth]{figures/rq2/f1micro_stagger_remove_classes.pdf}
        \includegraphics[width=0.24\textwidth]{figures/rq2/f1micro_stagger_add_implication.pdf}
        \includegraphics[width=0.24\textwidth]{figures/rq2/f1micro_stagger_remove_implication.pdf} \\
        
        \includegraphics[width=0.24\textwidth]{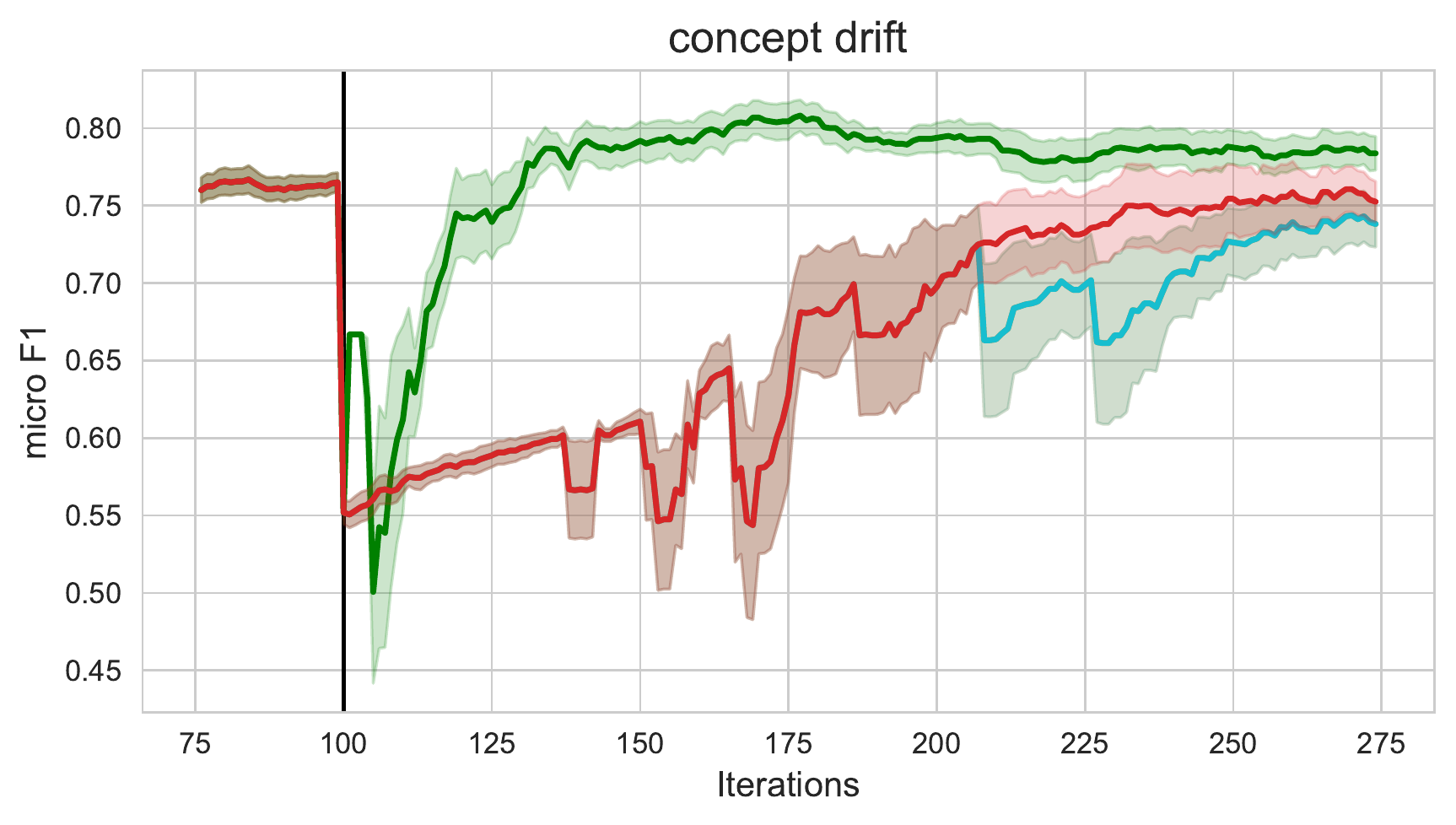}
        \includegraphics[width=0.24\textwidth]{figures/rq2/f1micro_emnist_remove_classes.pdf}
        \includegraphics[width=0.24\textwidth]{figures/rq2/f1micro_emnist_add_implication.pdf}
        \includegraphics[width=0.24\textwidth]{figures/rq2/f1micro_emnist_remove_implication.pdf} \\
        
        \includegraphics[width=0.24\textwidth]{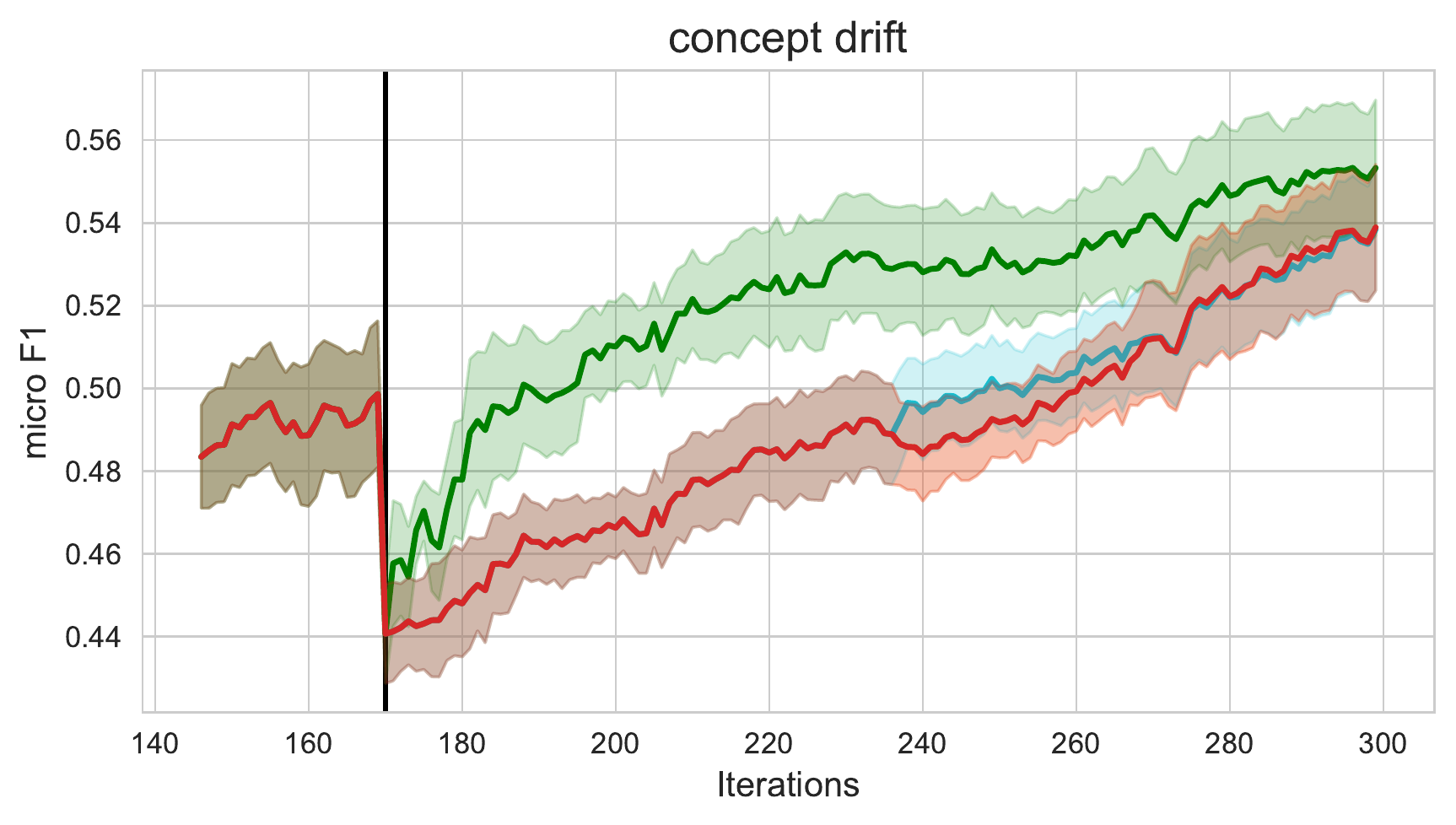}
        \includegraphics[width=0.24\textwidth]{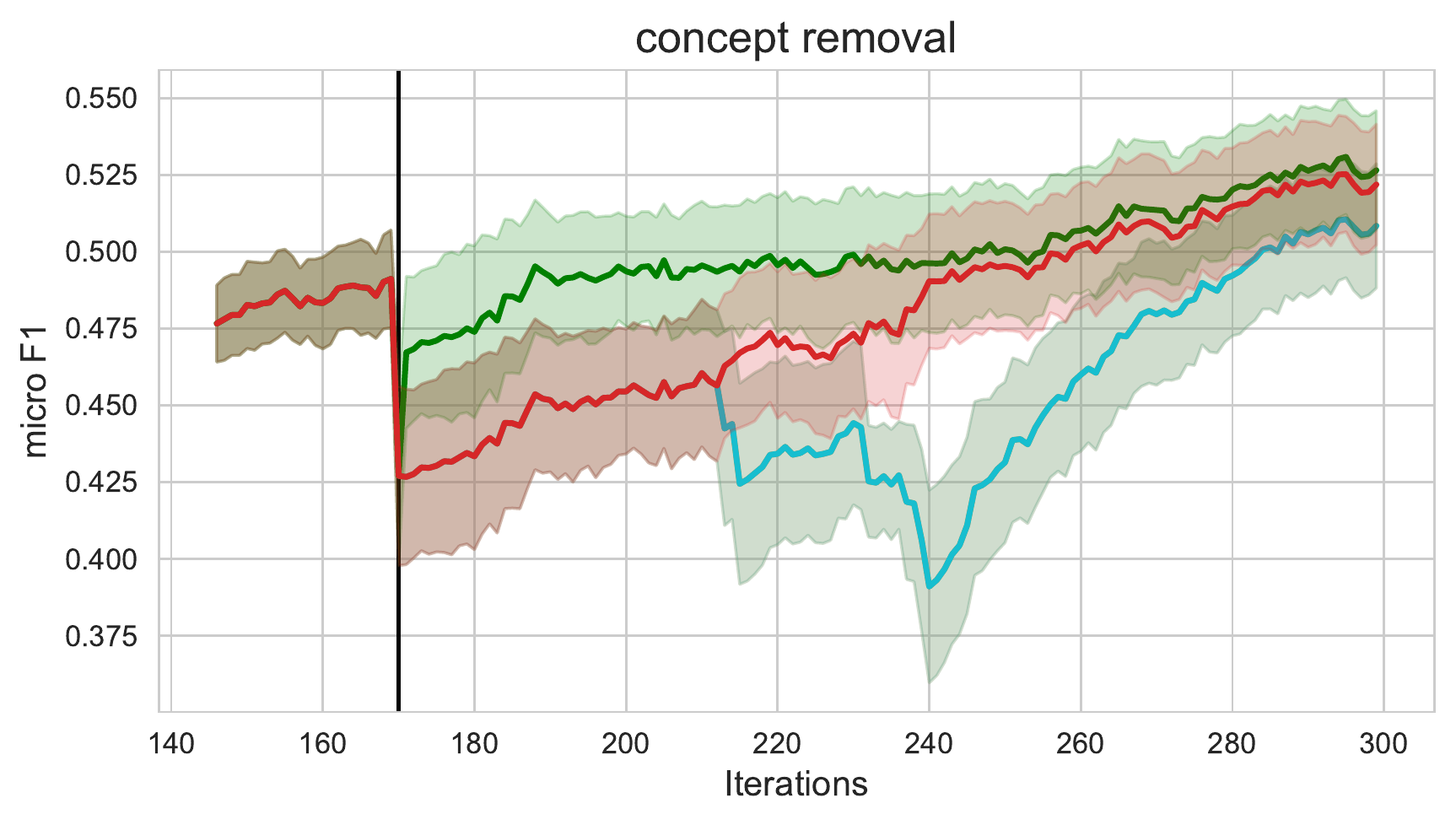}
        \includegraphics[width=0.24\textwidth]{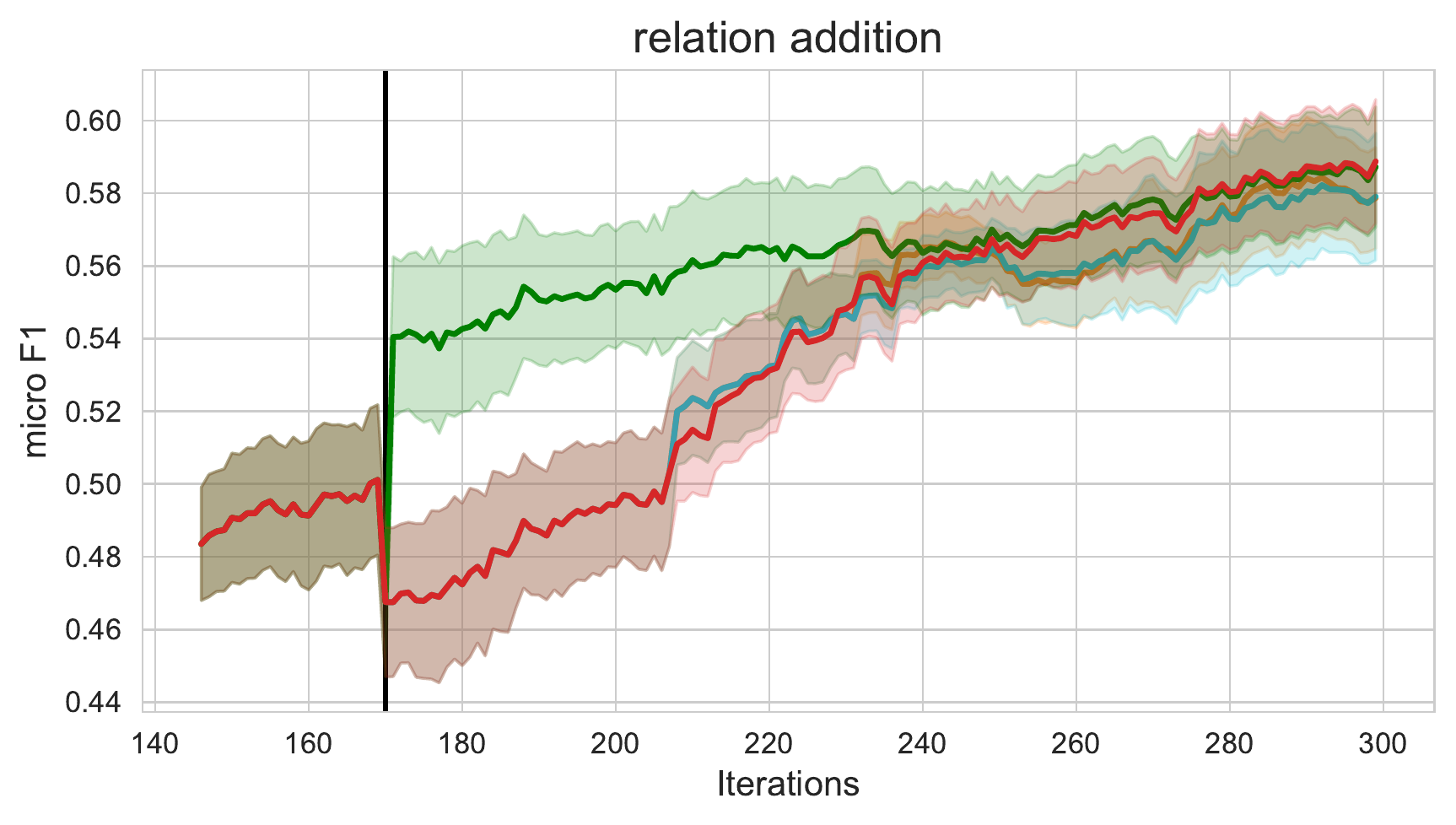}
        \includegraphics[width=0.24\textwidth]{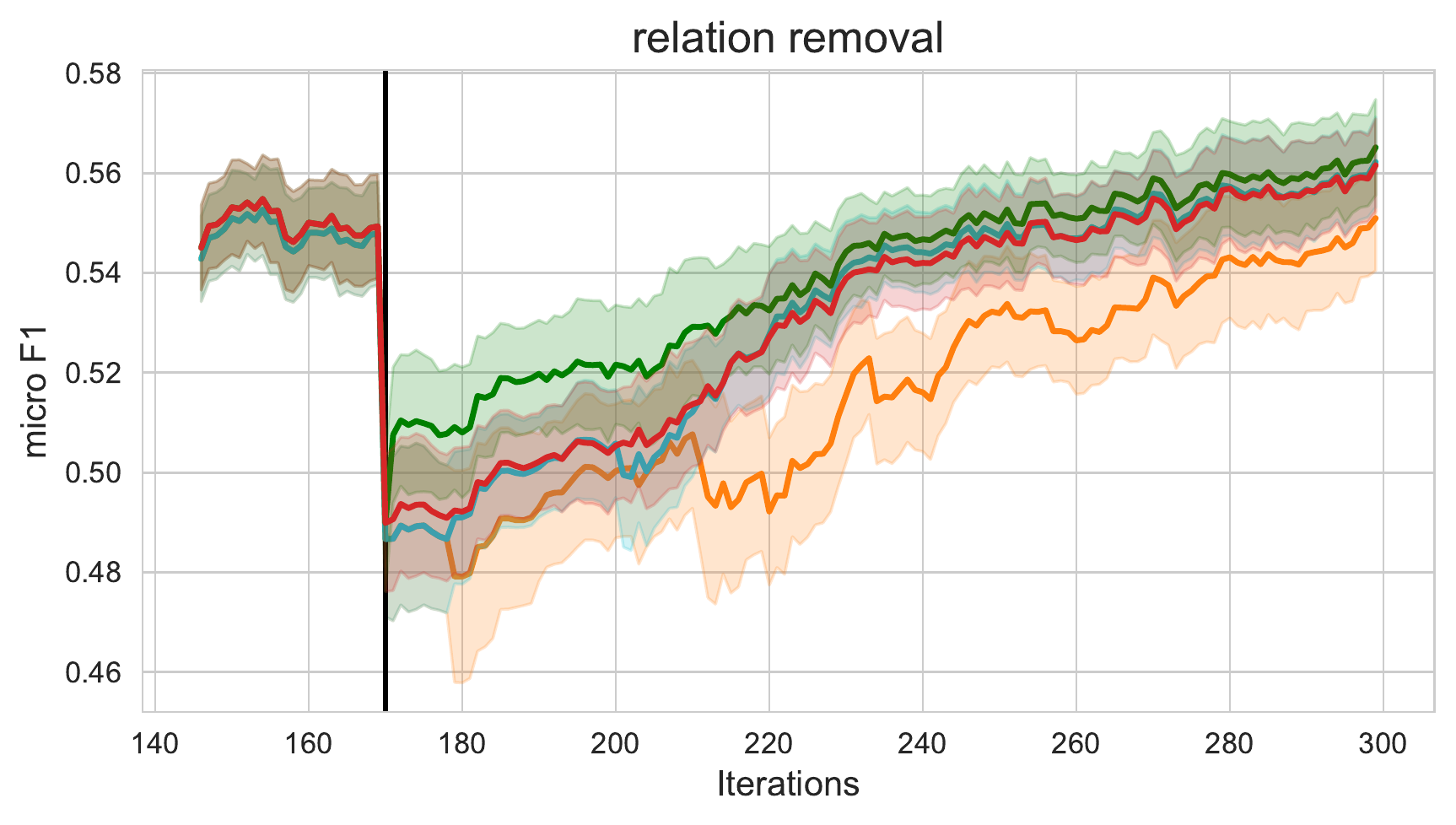} \\
    
    \end{tabular}
    \caption{Comparison in terms of micro F$_1$ between \acronym and less interactive variants.  Top to bottom:  results for HSTAGGER, EMNIST and 20NG.  Left to right:  concept drift, concept removal, relation addition, and relation removal.}
    \label{fig:q2f1}
\end{figure*}

The results in terms of micro $F_1$ score are reported in Figure~\ref{fig:q2f1}.  The ideal baseline outperforms all alternatives in all cases.  \acronym performs as well or better than the less interactive variants in all cases. The concept drift case for 20NG (first column, last plot), the momentary decrease in the performance around iteration 240 is due to the fact that in most of the runs MMD detects the drift and \acronym performs the adaptation around the same iteration.

\end{document}